\documentclass{sigkddExp}

\usepackage{hyperref}
\usepackage{url}

\usepackage{amsmath}
\usepackage{booktabs}

\usepackage{times}
\usepackage{latexsym}
\usepackage{graphicx}

\usepackage{multirow}
\usepackage{longtable}

\usepackage{wrapfig}

\usepackage[most]{tcolorbox}%it
\usepackage{enumitem}
\usepackage{subcaption}
\usepackage{lscape}
\usepackage{tabularx}
\usepackage{arydshln} % For dashed lines
\usepackage{natbib} 

\usepackage{threeparttable}

\usepackage{tikz}
\usetikzlibrary{positioning,arrows.meta,calc,decorations.pathreplacing}
\newtcolorbox[auto counter,number within=subsection]{myBox}[3][]{
arc=1mm,
lower separated=false,
fonttitle=\bfseries,
colbacktitle=gray!30, % Light gray title background
coltitle=black!20!black,   % Darker text for contrast
enhanced,
attach boxed title to top left={xshift=1mm,
        yshift=-1mm},
colframe=black!20!black,    % Darker border for better contrast
colback=cyan!10,%gray!20,      % Light gray background instead of white,
}

\newtcolorbox[auto counter,number within=subsection]{myfocus}[3][]{
arc=1mm,
lower separated=false,
fonttitle=\bfseries,
colbacktitle=gray!30, % Light gray title background
coltitle=black!20!black,   % Darker text for contrast
enhanced,
attach boxed title to top left={xshift=1mm,
        yshift=-1mm},
colframe=black!20!black,    % Darker border for better contrast
colback=red!10%gray!20,      % Light gray background instead of white
}

\usepackage{inconsolata}

\usepackage{tikz}

\usepackage{forest}
\usetikzlibrary{arrows.meta}
\begin{document}

\citestyle{acmauthoryear}

%
% --- Author Metadata here ---
% -- Can be completely blank or contain 'commented' information like this...
%\conferenceinfo{WOODSTOCK}{'97 El Paso, Texas USA} % If you happen to know the conference location etc.
%\CopyrightYear{2001} % Allows a non-default  copyright year  to be 'entered' - IF NEED BE.
%\crdata{0-12345-67-8/90/01}  % Allows non-default copyright data to be 'entered' - IF NEED BE.
% --- End of author Metadata ---

% \title{Unveiling Topological Structures from
% Language: \\A Survey of Topological Data
% Analysis Applications in Natural Language Processing}

% \title{Topological Data Analysis Applications in Natural Language Processing: A Survey}

% \title{A Survey on Topological Data Analysis Applications in Natural Language Processing}
\title{Topological Data Analysis Applications in Natural Language Processing: A Survey}

%\subtitle{[Extended Abstract]
% You need the command \numberofauthors to handle the "boxing"
% and alignment of the authors under the title, and to add
% a section for authors number 4 through n.
%
% Up to the first three authors are aligned under the title;
% use the \alignauthor commands below to handle those names
% and affiliations. Add names, affiliations, addresses for
% additional authors as the argument to \additionalauthors;
% these will be set for you without further effort on your
% part as the last section in the body of your article BEFORE
% References or any Appendices.

\numberofauthors{5}
%
% You can go ahead and credit authors number 4+ here;
% their names will appear in a section called
% "Additional Authors" just before the Appendices
% (if there are any) or Bibliography (if there
% aren't)

% Put no more than the first THREE authors in the \author command
%%You are free to format the authors in alternate ways if you have more 
%%than three authors.

% \author{

% % e-mail address with \email.
% \alignauthor Ben Trovato \\
%        \affaddr{Institute for Clarity in Documentation}\\
%        \affaddr{1932 Wallamaloo Lane}\\
%        \affaddr{Wallamaloo, New Zealand}\\
%        \email{trovato@corporation.com}
% \alignauthor G.K.M. Tobin\\
%        \affaddr{Institute for Clarity in Documentation}\\
%        \affaddr{P.O. Box 1212}\\
%        \affaddr{Dublin, Ohio 43017-6221}\\
%        \email{webmaster@marysville-ohio.com}
% \alignauthor Lars Th{\o}rv\"{a}ld\titlenote{This author is the
% one who did all the really hard work.}\\
%        \affaddr{The Th{\o}rv\"{a}ld Group}\\
%        \affaddr{1 Th{\o}rv\"{a}ld Circle}\\
%        \affaddr{Hekla, Iceland}\\
%        \email{larst@affiliation.org}
% }
% \additionalauthors{Additional authors: John Smith (The Th{\o}rvald Group,
% email: {\texttt{jsmith@affiliation.org}}) and Julius P.~Kumquat
% (The Kumquat Consortium, email: {\texttt{jpkumquat@consortium.net}}).}
% \date{30 July 1999}

\author{
%% e-mail address with \email.
\alignauthor Adaku Uchendu \\
       \affaddr{MIT Lincoln Laboratory}\\
       \affaddr{MA, USA}\\
%       \affaddr{USA}\\
       \email{adaku.uchendu@ll.mit.edu}
\alignauthor Thai Le\\
       \affaddr{Indiana University}\\
       \affaddr{IN, USA}\\
       %\affaddr{USA}\\
       \email{tle@iu.edu}
}

\maketitle

\begin{abstract}
% The surge of Internet-scale data has led to the adoption of diverse computational methods for extracting useful knowledge from complex datasets. Among these, Machine Learning (ML) has thrived as a powerful framework for pattern discovery, prediction, and representation learning. However, real-world data are often noisy, imbalanced, sparsely labeled, and high-dimensional, which motivates complementary perspectives for understanding their underlying structure. Topological Data Analysis (TDA) provides one such perspective by focusing on the intrinsic shape of data. Rather than serving as an alternative to ML, TDA can complement ML by capturing global structural and geometric properties that standard learning pipelines may not explicitly model. This has led researchers to explore TDA in conjunction with ML, especially for tasks in which structure is central to the data. Despite its promise, TDA has seen more limited adoption in Natural Language Processing (NLP) than in areas such as computer vision. Still, a dedicated research community has explored this direction, producing \textbf{130 papers} that we survey in this work. We categorize these studies into theoretical and non-theoretical approaches: the former use topology to explain linguistic phenomena, while the latter utilizing TDA on top of ML features based on diverse numerical representations. We conclude by outlining the main challenges and open questions in this emerging field. Resources and a curated list of papers are available at:
% \url{https://github.com/AdaUchendu/AwesomeTDA4NLP}
The surge of data available on the Internet has driven the adoption of a wide range of computational methods for analyzing and extracting insights from large-scale data. Among these, Machine Learning (ML) has become a central paradigm, offering powerful tools for pattern discovery, prediction, and representation learning across many domains. At the same time, real-world data often exhibit properties such as noise, imbalance, sparsity, limited supervision, and high dimensionality, motivating the use of additional analytical perspectives that can complement standard ML pipelines. One such perspective is Topological Data Analysis (TDA), a statistical framework that focuses on the intrinsic shape and structural organization of data. 
Rather than replacing ML, TDA offers a complementary lens for characterizing geometric and topological properties that may be difficult to capture with conventional feature-based or purely predictive approaches. 
This has motivated a growing body of work that integrates TDA into ML workflows, particularly in settings where data structure plays an important role. Despite this promise, TDA has received relatively limited attention in Natural Language Processing (NLP) compared to domains with more overt structural regularities, such as computer vision. Nevertheless, a dedicated community of researchers has explored its use in NLP, leading to \textbf{137 papers} that we comprehensively survey in this work. We organize these studies into theoretical and non-theoretical approaches. Theoretical approaches use topology to explain linguistic phenomena, whereas non-theoretical approaches incorporate TDA into ML-based pipelines through a variety of numerical representations. We conclude by discussing the key challenges and open questions that continue to shape this emerging area. Resources and a list of papers are available at:
\url{https://github.com/AdaUchendu/AwesomeTDA4NLP}\footnote{\small DISTRIBUTION STATEMENT A. Approved for public release. Distribution is unlimited.
This material is based upon work supported by the Department of the Air Force under Air Force Contract No. FA8702-15-D-0001 or FA8702-25-D-B002. Any opinions, findings, conclusions or recommendations expressed in this material are those of the author(s) and do not necessarily reflect the views of the Department of the Air Force.
© 2026 Massachusetts Institute of Technology.
Delivered to the U.S. Government with Unlimited Rights, as defined in DFARS Part 252.227-7013 or 7014 (Feb 2014). Notwithstanding any copyright notice, U.S. Government rights in this work are defined by DFARS 252.227-7013 or DFARS 252.227-7014 as detailed above. Use of this work other than as specifically authorized by the U.S. Government may violate any copyrights that exist in this work.}.
\end{abstract}

\section{Introduction}

\begin{figure}[tb!]
    \centering
    \includegraphics[width=1\linewidth]{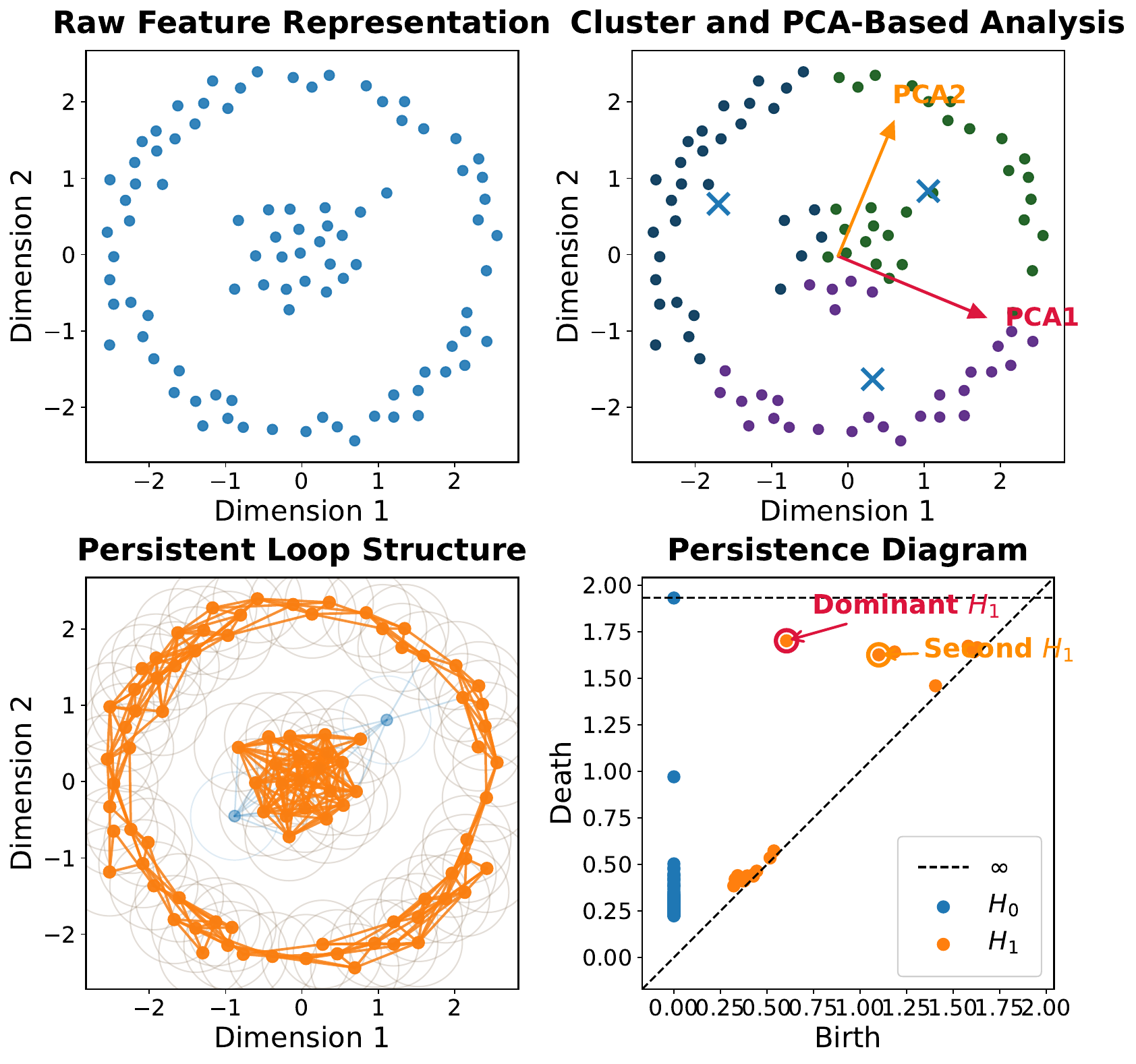}
    \caption{The raw point cloud contains a dense central region together with loop-like outer structure. Conventional ML analysis such as PCA and clustering is strongly influenced by variance and local density, so much of the summary is affected by the central cluster. In contrast, TDA emphasizes global structure. 
    In the persistence diagram, blue points represent \textit{connected component} features and orange points represent \textit{loop} features. For each topological feature, the horizontal coordinate (``birth'') is the scale at which the feature first appears, and the vertical coordinate (``death'') is the scale at which it disappears. Here, the prominent orange points that are further from the diagonal correspond to the persistent loop structures visible in the third panel.}
    \label{fig:teaser}
\end{figure}

Proliferation of the Internet has given rise to the generation of massive amounts of data. 
These massive amounts of data when processed can solve many crucial issues plaguing our current society. Due to this well-established notion among stake-holding institutions, the Machine Learning (ML) field has been thriving as a tool that extracts trends and solutions to non-trivial problems. 
However, real-world data tends to be noisy, heterogeneous, imbalanced, have missing labels, contain high-dimensionality, etc., often making the adoption of ML techniques to such datasets non-trivial. 
Therefore, to extract meaningful findings from data, specifically real-world data, clever techniques that extract additional features, while \textit{preserving the overall structure of the data} need to be employed. 

To that end, a small niche community for \textit{Topological Data Analysis (TDA) applications in NLP} has emerged. Being promised as a technique that can extract and analyze the shape/topology of data, TDA has great potential in mitigating such issues witnessed in real-world data. 
Thus, by applying TDA to NLP, we obtain ``\textit{topological structures from language},'' which 
refers not to intrinsic properties of raw text itself, but to the structures that emerge when linguistic data is mapped into high-dimensional embedding spaces. These induced topologies capture relationships among words, sentences, or documents based on their learned representations, rather than any inherent topological features of the text. 
%In addition, the phrase also helps us distinguish the application of TDA and other topological approaches applied to other fields, such as computer vision from NLP.  

% See Figure \ref{fig:teaser} for a visual illustration of how standard NLP techniques vs. TDA techniques capture properties from complex data. 

TDA is a ``collection of powerful tools that can quantify shape 
and structure in 
data''\footnote{\url{https://www.indicative.com/resource/topological-data-analysis/}} and is inspired by the algebraic topology 
and geometry mathematical fields. 
The benefits of TDA are vast, including the ability to extract additional features that are typically not captured by other feature extraction techniques~\cite{uchendu2023topformer,michel2017does,papamarkou2024position}. 
These features are known as \textit{topological features}.
%~\thai{is topological meaning closer to shape than structure? I think using either ``topological features'' or ``shape features'' would make the TDA topic more distinct from the reviewers' eyes}
Unsurprisingly, since TDA is used to capture topological features, it has been applied to many tasks where data has distinct graphical structures~\cite{papamarkou2024position,hensel2021survey}. 
These include tasks that have obvious graph-like structures, such as protein classification \cite{dey2018protein,lamine2023topological,valeriani2024geometry} and drug discovery \cite{alagappan2016multimodal}; to those that are not so obvious, such as diabetes classification \cite{wamil2023stratification,skaf2022topological}, 
image classification \cite{horntopological,trofimov2023learning}, and 
time series analysis \cite{lordgrilo,tymochko2021connections,gholizadeh2018short}. However, since the shape of a text is not apparent, it has not gained as much attention in NLP as it has in the Computer Vision field \cite{horntopological,trofimov2023learning} specifically in the Medical domains \cite{singh2023topological,nielson2015topological}. Still, several researchers have found ways to extract unique, global-level features using TDA. Other typical numerical representation techniques in text are unable to extract global-level features, making TDA suitable for the task. 
% signal processing \cite{wang2021topological},
% speech processing \cite{tulchinskii2023topological}.
% and other forms of graph prediction. 

Figure \ref{fig:teaser} illustrates a toy example that shows the utility of TDA to standard ML analysis. Conventional ML methods often summarize data through variance, centroids, or local grouping, which means dense regions can disproportionately shape the result. In the example, the central cluster absorbs much of the variance, while the outer loop structure plays a less visible role in PCA- and cluster-based summaries. TDA instead tracks topological features such as connected components and loops across multiple scales, allowing it to recover meaningful global structure. 
%Similar differences between TDA and variance-based ML analysis can also arise in many other distributions, especially when data contain loops, hollow regions, bridges, or other nontrivial geometric patterns.

% \begin{figure*}[!htb]
%     \centering
%     % \includegraphics[width=\linewidth]{figures/publications-2.pdf}
%     \includegraphics[width=0.8\linewidth]{publications_no_yaxis_name.pdf}
%     \caption{Number of TDA papers for NLP published each year from 2012 to March-2026}
%     \label{fig:evol}
% \end{figure*}

\begin{figure*}[!htb]
    \centering
    \includegraphics[width=0.95\linewidth]{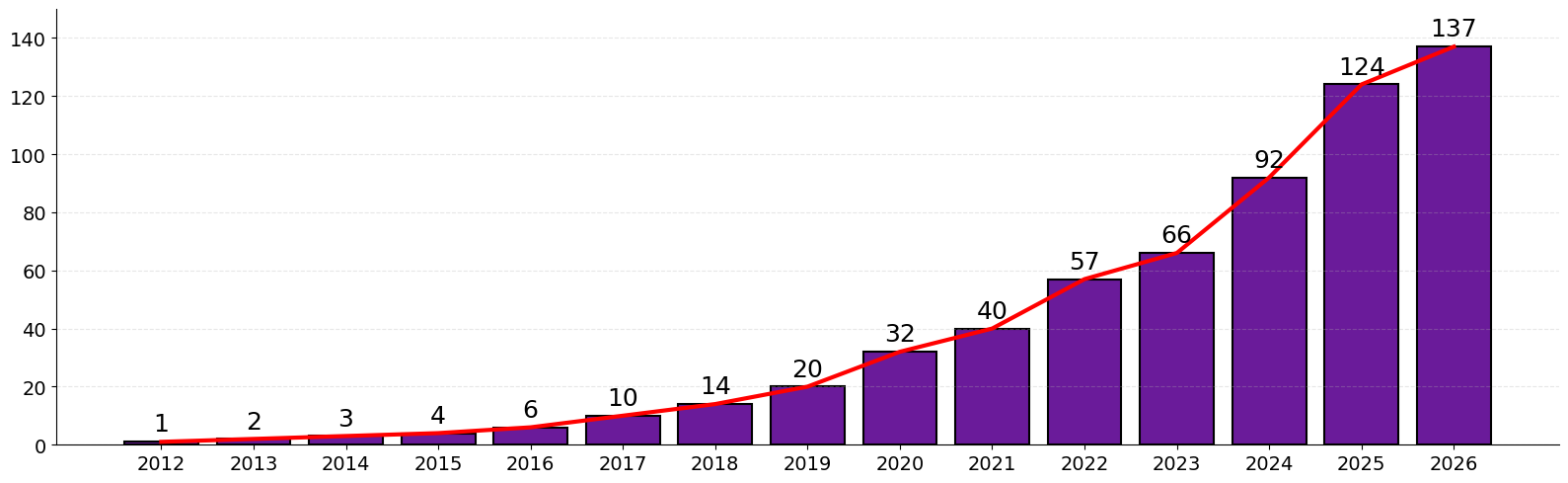} 
    % {publications_cumulative_trendline_biglabels.pdf} 
    %trendline
    % \includegraphics[width=0.95\linewidth]{publications_cumulative_no_redline_labels30.pdf} %notrendline
    
    % % \caption{Number of TDA papers for NLP published each year from 2012 to March-2026}
    \caption{Cumulative number of TDA papers for NLP published from 2012 to April-2026.}
    \label{fig:evol}
\end{figure*}

More broadly, TDA aims to answer the central research question - \textit{what is the true shape of data?} We survey \textbf{137 papers} that have attempted to find an answer through various approaches. 
% TDA's capacity to find the features after deformation techniques have been applied to the data makes TDA especially suitable for non-trivial NLP tasks. 
The first application of TDA in NLP was published in 2012 \cite{wagner2012computational}, and since then, there have been over 100 papers applying TDA in NLP. 
There have been a gradual acceleration in the number of published works in 
TDA applications on various NLP tasks, including ones pertaining to the recent emergence of Large Language Models (LLMs) such as 
hallucination detection~\cite{bazarova2025hallucination,samaga2026halluzig}, mechanistic interpretability~\cite{yan2025explainablemapperchartingllm,rathore2023topobert}, and model efficiency~\cite{gardinazzi2024persistent,more2025optimizing}; we project that this trend will continue in the future (Figure \ref{fig:evol}).
Therefore, based on these approaches, we
categorize these applications into two -  \textit{theoretical}~\cite{karlgren2014semantic,port2018persistent} and 
\textit{non-theoretical}~\cite{zhu2013persistent,doshi2018movie} approaches. 
\textit{Theoretical} approaches involve using TDA to explain linguistic phenomena by probing the topological space, shape, and evolution of topics. On the other hand, 
\textit{Non-theoretical} approaches mainly discuss how to effectively apply existing numerical representation techniques in NLP to extract novel topological features with TDA.

In addition, we observe that theoretical approaches only span 13 papers, while non-theoretical approaches have over 100 papers. Due to the higher number of non-theoretical approaches, we discuss several categories that could be useful for distinguishing applications: numerical representation, tasks, TDA technique, data modality, and learning type. 
TDA techniques (i.e., Persistent Homology and Mapper), data modality (i.e., text and speech), and learning type (i.e., unsupervised and supervised) are binary, making it difficult to meaningfully grasp distinctness from almost 120 papers. 
However, with tasks which refer to the problems in which approaches are adopted for; these have seven categories - \textit{(1) classification, (2) clustering \& topic modeling, (3) sentiment \& semantic analysis, (4) structure \& visualization, (5) health, social, \& scholarly analysis, (6) speech processing,} and \textit{(7) model interpretation \& analysis}. 
We observe that classification and model interpretation \& analysis are the most popular tasks explored by researchers. 
In addition, numerical representations leveraged in non-theoretical 
application include - \textit{(1) TF-IDF, (2) Word2Vec, (3) GloVe, (4) FastText, (5) ELMo, (6) Transformers, (7) Symbolic,} and \textit{(8) Multi-Modal}. We use numerical representation as our main taxonomy for non-theoretical applications because it is the bottleneck for extracting topological features from text. 

\begin{figure}[tb!]
    \centering
    \includegraphics[width=1\linewidth]{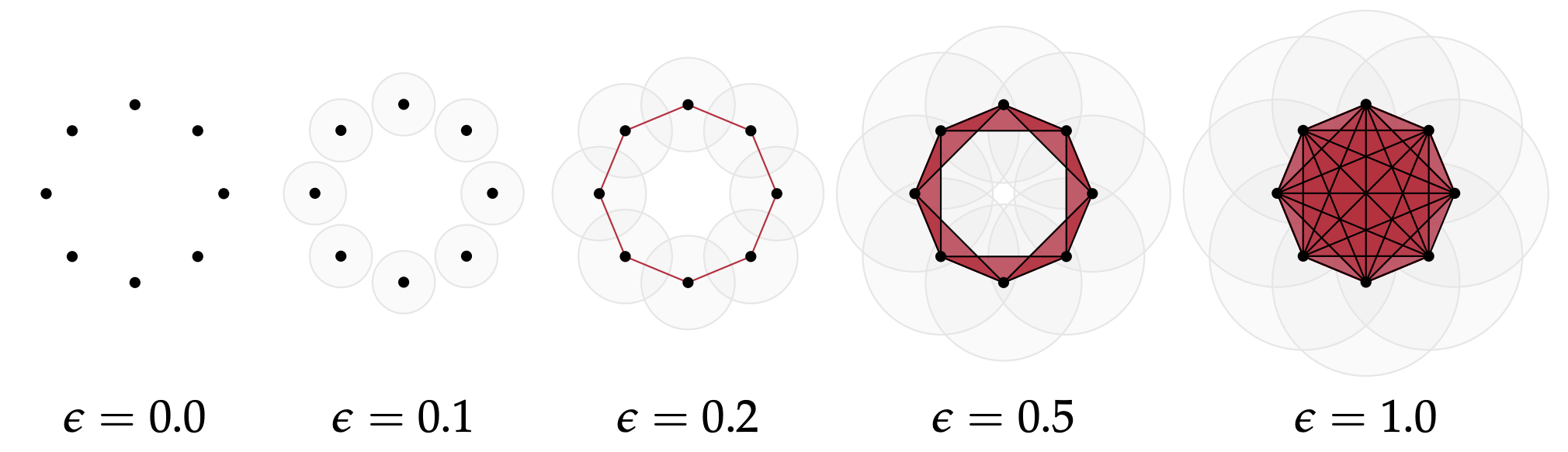}
    \caption{Illustration of the Persistent Homology technique using different radii to find the persistent features \cite{toptutorial2020}. $\epsilon$ is the ball diameter.}
    \label{fig:pers}
    % \vspace{-10pt}
\end{figure}

Finally, we will first discuss the principles behind TDA and the two main techniques employed for TDA feature extraction: \textit{Persistent Homology} and \textit{Mapper}.
In addition, we will discuss the selection criteria and taxonomy development of the survey, both the theoretical and non-theoretical approaches, and discuss interesting findings, open problems, and future directions.

\section{Topological Data Analysis}
Topology is defined as \textit{``the study of geometric properties 
and spatial relations unaffected by the continuous change 
of shape or size of figures,''} (Oxford Dictionary). 
TDA is then a collection of powerful techniques that can quantify the shape and structure 
of data~\cite{munch2017user}. 
Two main techniques are used to extract TDA features: \textit{Persistent Homology} and \textit{Mapper}. 
% ~\thai{This section is essential, but feel like just a little bit too technical. I think, if we can accompany Figure 2 and Figure 3 with one specific example in NLP, that would be more friendly and interesting. Like, this is the PH for this NLP task from this paper, and this is the Mapper for this NLP task or for this dataset in this paper.}

\subsection{Persistent Homology}
Persistent Homology (PH) \cite{edelsbrunner2008persistent} 
is the most popular TDA technique. It uses algebraic topology 
methods to extract topological signatures at different spatial dimensions. 
This process involves representing data as a point cloud and performing deformation or perturbation processes to extract the true ``shape'' 
of data after the noise has been removed. To achieve this, 
PH employs  
Vietoris-Rips complex \cite{munch2017user}. 
Vietoris-Rips complex is a way to build simplicial complexes which are used to 
represent data in a topological space. 
A simplicial complex is a topological space built by putting points, lines, and higher dimensional shapes together.  
These formations reveal features that are holes in different dimensions, 
represented as \textit{betti numbers} 
($\beta_d$, $d$-dimension). 
Holes in the 0-dimension ($\beta_0$) is represented as one vertex,  
1-dimension ($\beta_1$) is represented as an edge,  
2-dimension ($\beta_2$) is represented as a triangle. 
Further, these features are called connected components, loops/tunnels, and voids, respectively. 

% \begin{figure}[tb!]
%     \centering
%     \includegraphics[width=1\linewidth]{figures/vr_rieck.png}
%     \caption{Illustration of the Persistent Homology technique using different radii to find the persistent features \cite{toptutorial2020}. $\epsilon$ is the ball diameter.}
%     \label{fig:pers}
%     % \vspace{-10pt}
% \end{figure}

Using the method described above, data is represented as a point cloud, 
and circles are drawn around each point. Next, the radius of each circle is increased using a defined range of points, such that if the circles get bigger and touch, one of the points disappears and this is recorded as a $death$. Additionally, this process of perturbation in different dimensions can cause the $birth$ of a new hole, which is also recorded. 
%Thus, due to these deformations, the following TDA features can be extracted and recorded in a 3-column matrix, which consists of columns representing - the $birth$ (formation of holes), $death$ (deformation or the closing of holes), and persistence features. 
Persistence is defined as the length of time it took a feature to disappear or die ($death - birth$). The $death$ is recorded with the radius value at which the points 
overlap. Lastly, TDA features are typically visualized in a persistence diagram, which is a visual representation of the $birth$ (x-axis) vs. $death$ (y-axis) features. 
Other ways of visualizing TDA features include persistence images \cite{adams2017persistence} and barcode plots \cite{ghrist2008barcodes}.
Figure \ref{fig:pers} illustrates an example of the process of extracting TDA features using PH. 
In terms of application, PH has been used to extract novel features to complement existing NLP representations and improve various classification performances~\cite{doshi2018movie,uchendu2023topformer,wu2022topological}.

\begin{myfocus}[]{Definition}%{AA}
\textsc{\textbf{Persistent Homology.}} 
% \footnotesize
\textit{
This is a TDA technique that studies the deformation of ``holes'' 
in different dimensions. Using PH, we can track when features appear and disappear and visualize these features, usually
in a persistence diagram. This process allows us to find the true structure of data, typically devoid of noise. 
}
\end{myfocus}

% \begin{figure}[!htb]
%     \centering
%     \includegraphics[width=0.7 \linewidth]{figures/mapper_better.png}
%     \caption{Illustration of Mapper from \citet{murugan2019introduction}. The filter function $f$ is a height function, which is a projection onto the y-axis. The cover of the projected space is the four intervals $U_i$. The Mapper graph on the right is a result of applying the rest of the Mapper algorithm and clustering each preimage in the nearest neighbor.  }
%     \label{fig:mapper}
%     % \vspace{-10pt}
% \end{figure}

\subsection{Mapper}
\begin{figure}[tb!]
\centering
\includegraphics[width=0.9\linewidth]{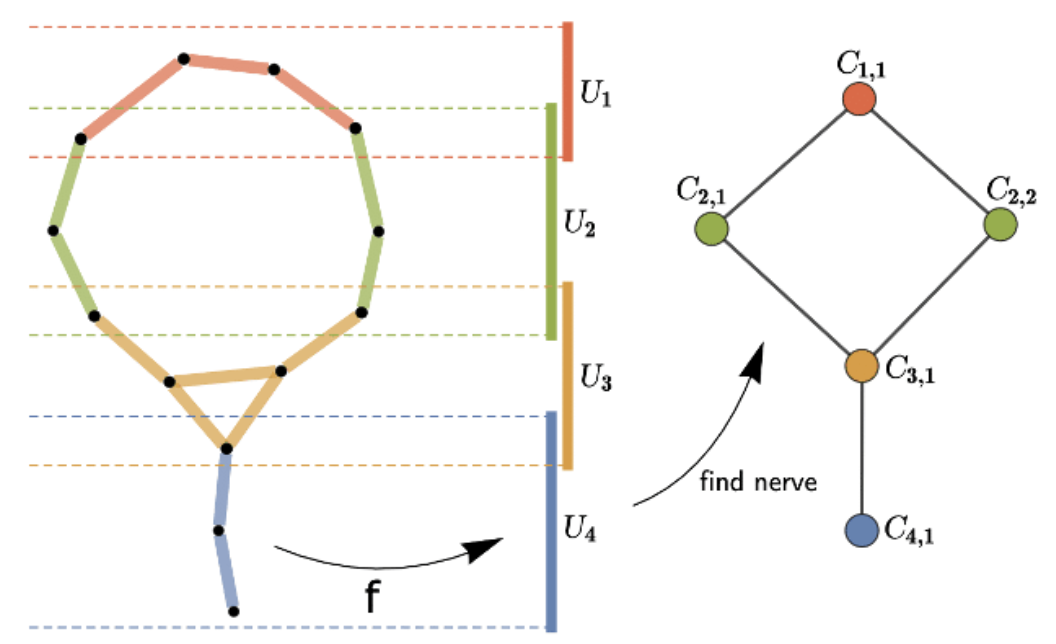}
\caption{Illustration of Mapper from \cite{murugan2019introduction}. The filter function $f$ is a height function, which is a projection onto the y-axis. The cover of the projected space is the four intervals $U_i$. The Mapper graph on the right is a result of applying the rest of the Mapper algorithm and clustering each preimage in the nearest neighbor.}
\label{fig:mapper}
\end{figure}

Mapper is a dimension reduction clustering technique for visualizing TDA-extracted topological structures/signatures. It was proposed by Singh et al. \cite{singh2007topological} and has been used extensively to visualize topological structures in data to create visually pleasing figures. In addition, Mapper figures have been used to interpret model performance through data probing \cite{carlsson2020topological}. 
The Mapper algorithm works in four steps\footnote{\url{https://www.quantmetry.com/blog/topological-data-analysis-with-mapper/}} (Figure \ref{fig:mapper}) following the instructions of \cite{murugan2019introduction}: 
(1) Transform the data to a lower-dimensional space using a filter function $f$, also known as a lens. This implies projecting from one space to another. 
Options for filter functions include PCA~\cite{mackiewicz1993principal}, UMAP~\cite{mcinnes2018umap}, and any other dimension-reduction algorithms; 
(2) Create a cover \((U_i)_{i\in I}\) for the projected space, which is typically composed of overlapping intervals with a constant length;
(3) Cluster the points in the preimage \( f^{-1}(U_i) \) into sets \( C_{i,1}, \ldots, C_{i,k_i} \) per interval \( U_i \);
(4) Create a graph where each vertex represents a cluster set. There is an edge between two vertices if the corresponding clusters share common points. 
Points in the same neighborhood are clustered using a defined clustering technique, such as DBSCAN~\cite{ester1996density} to change a cluster of several points into a node of a graph.

The intrinsic nature of the Mapper algorithm makes it advantageous in 
preserving structure, even with mapping from one dimension to another. 
Furthermore, the clustering techniques allow it to be used to explain model performance as the clusters and colors have meaning that can be further explored. 
% In this survey, we will discuss how several researchers used it to explain several phenomena in NLP tasks. 
% Finally, PH is focused on the quantitative analysis of topological features that persist across different scales, providing detailed insights into the global structure of data. 
% While, Mapper, is more of a qualitative, visual tool, which creates a graph-based representation of data based on a lens function. 
Finally, Mapper is more useful for exploratory data analysis, while PH is more useful for 
analyzing point clouds and examining the persistence of features. 
In this survey, we will discuss how several researchers use Mapper to explain or enhance several phenomena in NLP tasks.

\begin{myfocus}[]{Definition}%{AA}
\textsc{\textbf{Mapper.}} 
% \footnotesize
\textit{
This is a TDA technique that visualizes the graphical representation of data in order to capture the 
intrinsic structure. It is very useful for preserving data structure and creating visually pleasing plots, which can be 
investigated manually to find insights. 
}
\end{myfocus}

%~\thai{I feel like there should be a sentence describing the distinct difference between Mapper and PH.}

\begin{figure*}[!htb]
    \centering
    \includegraphics[width=1\linewidth]{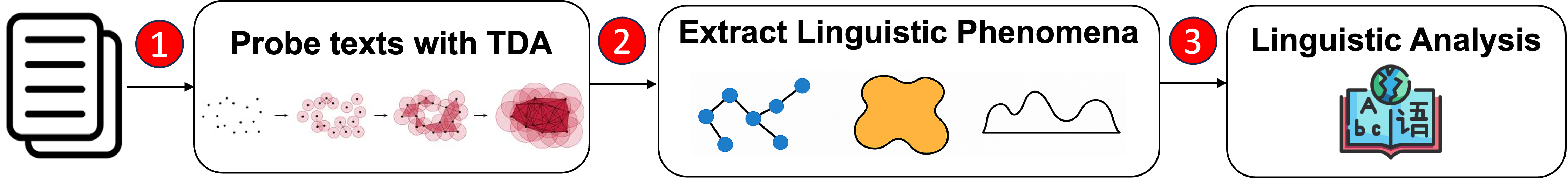}
    \caption{Illustration of the theoretical approaches researchers have employed to (1) probe texts, (2) extract TDA features, (3) use these features to explain or confirm known linguistic phenomena.}
    \label{fig:theo_pipeline}
\end{figure*}

\begin{figure*}[!htb]
    \centering
    \includegraphics[width=1\linewidth]{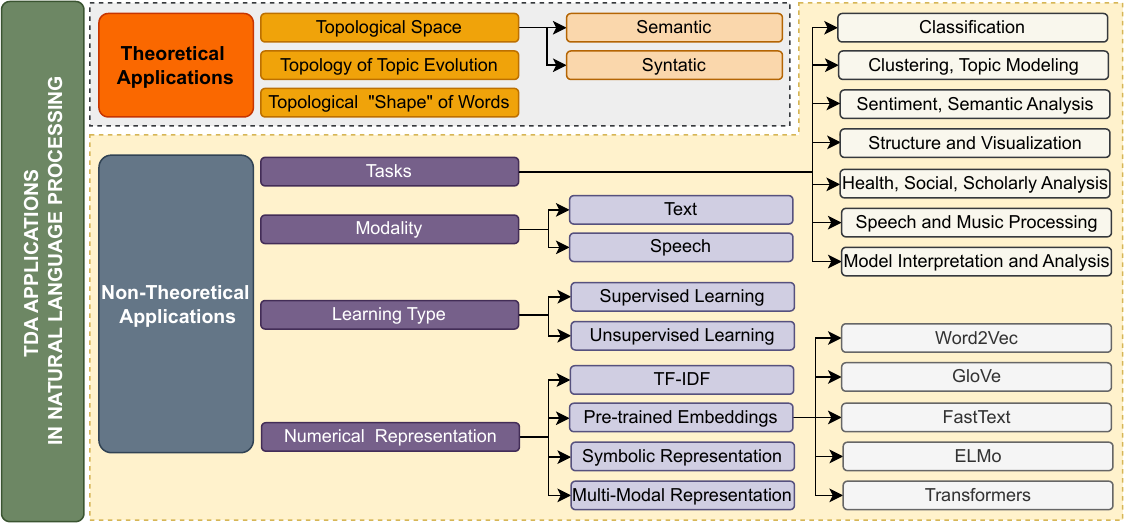}
    \caption{Taxonomy of Topological Data Analysis (TDA) for Natural Language Processing (NLP) Applications}
    \label{fig:flowchart}
    % \vspace{-10pt}
\end{figure*}

\section{Survey Scope}

\subsection{Selection Criteria}
In order to find all NLP papers that applied TDA, we manually searched on Google Scholar using key terms such as 
\textit{text mining persistent homology, language model topological data analysis}, etc., checking related articles of the relevant papers, their cited papers, and different combinations of all three methods. 
After, obtaining over 60 papers initially, we started creating a taxonomy and categorizing the papers. 
Initially, we focused on TDA applications in textual data, but as we searched, we found several applications in speech, and collected such papers. Finally, we removed papers that did not apply TDA to text or human speech data. 
Papers removed, either applied TDA to a graphical representation of reddit social networks, applied non-TDA topological techniques, or 
applied TDA to non-speech audio data. 
Using these criteria we selected only papers that fit schema and collected the rest following the same schema.

\subsection{Taxonomy Development}
Based on the papers selected for the survey, we were able to categorize the applications of these papers into two approaches - 
\textit{theoretical} and \textit{non-theoretical} applications:

\noindent \textbf{Theoretical applications of TDA in NLP}: These focus on understanding, characterizing, or proving properties of language and its representations through the lens of topology. They are less about immediate performance gains and more about insight. This application aims to answer the question - ``What do the shapes of embedding spaces tell us about language itself and our models of it?'' Example - Analyzing embedding spaces: Using persistent homology to study whether semantic clusters, or syntactic structures, correspond to stable topological features, and finding out what that tells us about language.

\noindent \textbf{Non-theoretical (practical) applications of TDA in NLP}: These treat TDA as a tool for solving tasks, regardless of whether deeper linguistic/topological insights are obtained. The emphasis is on utility. This application aims to answer the question - ``How can topological summaries directly help with applied NLP tasks?'' Example - Feature engineering: Augmenting classifiers for sentiment analysis, topic detection, or authorship attribution with topological signatures.

% \begin{itemize}[leftmargin=\dimexpr\parindent+0.8\labelwidth\relax,itemsep=3pt]
%     % \item \textbf{Theoretical applications of TDA in NLP}: These focus on understanding, characterizing, or proving properties of language and its representations through the lens of topology. They are less about immediate performance gains and more about insight. This application aims to answer the question - ``What do the shapes of embedding spaces tell us about language itself and our models of it?'' Example - Analyzing embedding spaces: Using persistent homology to study whether semantic clusters, or syntactic structures, correspond to stable topological features, and finding out what that tells us about language.

%     \item \textbf{Non-theoretical (practical) applications of TDA in NLP}: These treat TDA as a tool for solving tasks, regardless of whether deeper linguistic/topological insights are obtained. The emphasis is on utility. This application aims to answer the question - ``How can topological summaries directly help with applied NLP tasks?'' Example - Feature engineering: Augmenting classifiers for sentiment analysis, topic detection, or authorship attribution with topological signatures.
% \end{itemize}

\begin{table*}[!htb]
    \centering
    \caption{Theoretical Applications of TDA in NLP}
    \begin{tabularx}{\textwidth}{l l X l}
        \toprule
        \textbf{Name} & \textbf{Category} & \textbf{Task} & \textbf{TDA Technique} \\
        \midrule

        \cite{karlgren2014semantic} & TS-Sem & Identify topical density of the space & Mapper \\
        % \hdashline
        \cite{cavaliere2017context} & TS-Sem & Extracts main concepts from text & Persistent Homology \\
        % \hdashline
        \cite{wagner2012computational} & TS-Sem & Analyzes document similarities & Persistent Homology \\
        % \hdashline
        \cite{gromov2025language} & TS-Sem & Analyzes holes in English and Russian languages & Persistent Homology \\
        % \hdashline 
        \cite{Sakib2025AffixIsometry} & TS-Sem & Measures how word affixes (i.e., -ness) distort or preserve semantic structure & Persistent Homology \\
        % \hdashline
        \cite{christianson2020architecture} & TS-Sem & Creates a topological structure for college-level linear algebra texts & Persistent Homology \\

         \cite{port2018persistent} & TS-Syn & Analyzes syntactic parameters of different language families & Persistent Homology \\
        % \hdashline
        \cite{port2019topological} & TS-Syn & Explains linguistic structures with homoplasy phenomena & Persistent Homology \\
        % \hline

        % \hline
        
        \cite{draganov2024shape} & TSW & Investigates the ``shape'' of language phylogenies in the Indo-European language family & Persistent Homology \\
        % \hdashline
        \cite{fitz2022shape} & TSW & Captures grammatical structure expressed by corpus using \textit{word manifold} & Persistent Homology \\
        % \hdashline
        \cite{dong2024linguistics} & TSW & Analyzes shapes of South American languages: Nuclear-Macro-J\^{e} \& Quechuan families & Persistent Homology \\
        % \hdashline
        \cite{bouazzaoui2021application} & TSW & Analyzes topological similarity between Tifinagh and Phoenician scripts & Persistent Homology \\
        % \hline

        \cite{sami2017simplified} & TTE & Topic evolution within documents & Persistent Homology \\
        \bottomrule
    \end{tabularx}
    
    \vspace{2pt}
    \begin{minipage}{\textwidth}
    \footnotesize
    \textit{Note:} TS-Sem = Topological Space (Semantic); TS-Syn = Topological Space (Syntactic); TTE = Topology of Topic Evolution; TSW = Topological ``Shape'' of Words.
    \end{minipage}
    
    \label{tab:theory_label}
\end{table*}

\section{Theoretical Approaches}
Since the field of NLP is very interested in representing and analyzing texts or speech 
in meaningful ways, several theoretical approaches have been proposed 
to investigate how well 
these approaches align with linguistic principles. 
Thus to explain or confirm linguistic phenomena within the NLP paradigm, 
a few researchers have proposed topological approaches for probing data. 
See Figure \ref{fig:theo_pipeline} for an illustration of this pipeline. 
By employing TDA techniques - Persistent homology or Mapper to 
probe for linguistic phenomena, researchers aim to 
capture  the \textit{topological space} (both \textit{semantic} and \textit{syntactic} relationships) in language,  
analyze and visualize the \textit{topology of topic evolution} within texts, 
and extract the \textit{topological shape} of words. 
See Table \ref{tab:theory_label} for the theoretical approaches and Figure \ref{fig:flowchart} for the flowchart illustrating the taxonomy of theoretical 
applications of TDA in NLP tasks. 
In essence, these theoretical topological methods provide a conceptual bridge between linguistic theory and mathematical topology.

% \begin{myfocus}[]{Definition}%{AA}
% \textsc{\textbf{Theoretical vs. Non-Theoretical approaches.}} 
% % \footnotesize
% \textit{The main difference between the theoretical and non-theoretical approaches in this context is - 
% \underline{Theoretical approaches} use TDA (sometimes in combination with other numerical techniques) to understand and explain linguistic phenomena, 
% while \underline{Non-theoretical approaches} use TDA to enhance or explain model performance. 
% This means that the {Theoretical approaches} are focused on linguistic phenomena, while {Non-theoretical approaches} are
% focused on model performance. 
% Thus, using linguistic phenomena understood from the 
% {Theoretical approaches}, can inform and enhance {Non-theoretical approaches}. 
% }
% \end{myfocus}

\subsection{Topological Space}

\subsubsection{Semantic Topological Space} \label{sec:semantic}
A \textbf{semantic topological space} is a conceptual framework used to 
represent and analyze the relationships between the meanings (semantics) of words, 
phrases, or other linguistic units in a topological or shape structure. This representation 
involves mapping these units into a mathematical space where the distance or structure 
between them reflects semantic similarity or other relationships (i.e., Euclidean space $\to$ Topological space).

% Due to the limitations of traditional clustering algorithms in capturing semantic relationships within texts, \citet{chiang2007discover} utilizes TDA and represents semantic space 
% as a simplicial complex of Euclidean space. 
% This technique captures the coherent topics of a document 
% in the connected component dimension ($\beta_0$), representing the semantic topological space \cite{chiang2007discover}.
Karlgren et al. \cite{karlgren2014semantic} 
visualizes the topological semantic space of text using Mapper, which identifies the topical density of the space. 
To capture topological properties,
they train two semantic spaces in a specific topical domain \cite{karlgren2014semantic}. 
One space was trained only on articles of similar topics, and 
the other on introductory paragraphs of those same articles. 
Findings reveal that clusters of main concepts remained close for the space trained only on articles of similar topics. 
For the other topological space, the main concepts were randomly distributed \cite{karlgren2014semantic}. 
This suggests that semantic topological space can be better 
captured with richer and denser data than with sparser data.

In addition, Cavaliere et al. \cite{cavaliere2017context} extracts main concepts from the texts by
probing the context-aware semantic topological space built with simplicial complexes.
Gromov et al. \cite{gromov2025language} builds a semantic space with bigrams and trigrams of Word2vec embeddings of English and Russian languages to ascertain how distinct the languages are. Next, they use these findings for bot detection.
Sakib et al. \cite{Sakib2025AffixIsometry} 
proposes a metric - \textit{S M Nazmuz Sakib Topological Affix Isometry Index} 
to measure the structural preservation of the semantic space after the addition 
of affixes (i.e., -ness, -er, un-). 
They use persistent homology to create word manifolds to measure how affixes  
preserve (i.e., isometry) or distorts 
the meanings and relationships of the words it attaches to. 
Christianson et al. \cite{christianson2020architecture} uses persistent homology to structure the semantic networks 
from mathematical concepts in 
college-level linear algebra texts. 
They find that the networks show strong core-periphery architecture, where concepts are dense and sparse periphery for concepts presented throughout. 
Their results could inform the optimal design of 
principles for textbooks.

Furthermore, 
Wagner et al. \cite{wagner2012computational} uses TF-IDF to numerically represent the top 10-50 words in a corpus and build a topological space that analyzes the structure of similarities within several documents. This topological space is built 
using discrete Morse theory and persistent homology to find meaningful topological patterns \cite{wagner2012computational}. 
However, in 2012, they found that their technique was unsuccessful due to computational costs, which is a testament to how the field of NLP has improved so that we now have more tractable solutions, such as 
dimensionality reduction algorithms \cite{mcinnes2018umap}, and 
TDA packages (Ripser \cite{bauer2021ripser}, Sklearn-TDA \cite{scikittda2019}, PHAT \cite{bauer2017phat}, pytorch-topological\footnote{\url{https://github.com/aidos-lab/pytorch-topological}}), as well as compute resources
to efficiently construct topological spaces from large or complex data.

\begin{myBox}[]{Definition}%{AA}
\textsc{\textbf{Insight (Semantic Space).}} 
% \footnotesize
\textit{The semantic topological space is explored by researchers to identify semantic linguistic principles captured in texts through a topological lens. 
Most of the applications in this section involve understanding the 
semantic similarity between text pairs from a topological lens. 
}
\end{myBox}

\subsubsection{Syntactic Topological Space}
A \textbf{syntactic topological space} is a theoretical framework used to represent and analyze the relationships between syntactic structures from a topological perspective. 
This concept is particularly relevant in linguistics, where it helps model and understand the structural aspects of language, such as grammar, sentence construction, or the hierarchical organization of those linguistic units.

Therefore, Port et al. \cite{port2018persistent} analyzes how syntactic parameters are distributed 
over different language families, including Indo-European, Niger-Congo, Austronesian, and Afro-Asiatic families. 
For instance, features in $\beta_0$ capture the subdivision into historical, and features in $\beta_1$ capture syntactic differences between branches of families of languages, as well as the syntactic influences between them \cite{port2018persistent}.
They investigate the syntactic topological structures of language families, 
specifically Indo-European, Niger-Congo, Austronesian, and Afro-Asiatic families.
Port et al. 
\cite{port2018persistent} shows that the three persistent connected components ($\beta_0$) 
in the Niger-Congo family represents its
three subfamilies - Mande, Atlantic-Congo, and Kordofania.
The syntactic topological structures of these languages also
reveal the historical linguistic phenomena that the Hellenic branch played
a role in the historical development of the Indo-European languages \cite{port2018persistent}. 

Similarly, Port et al. \cite{port2019topological} probes the 
interpretability of the syntactic topological space by 
introducing \textit{homoplasy} phenomena to explain persistent loops. 
Homoplasy phenomena in syntax are observed when dissimilar languages exhibit
syntactic similarities \cite{port2019topological}. 
Findings reveal that the Indo-European family languages - Czech, Lithuanian, Middle Dutch, and Swiss German have 
the same homoplasy phenomena \cite{port2019topological} due 
to the appearance of persistent loops in these languages.
This is because Middle Dutch and Swiss German are similar, while Czech and Lithuanian are so different from them, making the homoplasy phenomenon the most reasonable explanation \cite{port2019topological}. 

\begin{myBox}[]{Definition}%{AA}
\textsc{\textbf{Insight (Syntactic Space).}} 
% \footnotesize
\textit{The syntactic topological space captures the syntactic structure of language (i.e., grammar, etc.) from a topological lens. 
Using this framework, researchers confirm linguistic phenomena in language families and subfamilies by exploring the syntactic relationship between 
languages. 
Thus, a novel application of this framework could include the discovery of 
new linguistic phenomena within syntactic structures.
}
\end{myBox}

\subsection{Topology of Topic Evolution}
The \textbf{topology of topic evolution} refers to the study and representation of how topics, themes, or concepts develop and change over time within a given corpus of texts or discourses in a topological space/structure. 
This concept is particularly relevant in fields where understanding the temporal dynamics of topics can provide insights into trends, shifts in public opinion, or the development of scientific or cultural themes.

Sami et al. \cite{sami2017simplified} utilizes TDA to visualize the relationship between words in a text block, words in a corpus, and text blocks in a corpus. 
Text blocks represent a chapter/section in a book, a document in a media corpus, and a webpage in a web corpus \cite{sami2017simplified}. 
They visualize both local context (i.e., each text block in a set of sentences) and global context (i.e., occurrence of extracted words in the corpus) features. 
These features are extracted by using the circular topology to represent words.
Then, the peripheral nature of the text block and corpus can be visualized using these features. 
With the Local context features, dimension reduction is achieved by stemming the prefixes and suffixes of words. 
For the Global context features, word movement is captured, which analyzes topic evolution. 
Finally, findings reveal that using the circular topology in 2D space,  
core words from the corpus stay close to the center, 
and the explanatory words remain close to the circle's periphery.

\begin{myBox}[]{Definition}%{AA}
\textsc{\textbf{Insight (Topic Evolution).}} 
% \footnotesize
\textit{Exploring the topology of topic evolution is a novel framework for capturing the topology of topics in a corpus. The findings suggest that this framework can be adopted to evaluate the utility of a summarization, paraphrasing, or obfuscating model predictions by comparing the topology of the topic evolution in the original vs. the perturbed texts. 
}
\end{myBox}

\subsection{Topological ``Shape'' of Words}
The \textbf{topological ``shape'' of words} is a conceptual framework in linguistics and cognitive science that explores the structural properties of \textit{words}.  
This framework leverages ideas from topology to capture the true shape of words in a linguistically meaningful way. 
Thus, using topological methods such as TDA, 
the structural properties of words can be extracted and analyzed.

Draganov et al. \cite{draganov2024shape} captures the ``shape'' of words for several languages by 
comparing the phylogenies or evolutionary history of language
in the Indo-European language family.
Initially, numerically representing the texts with FastText \cite{bojanowski2017enriching}, they 
use persistent homology to construct language phylogenetic trees
for over 81 Indo-European languages. 
Experiments reveal that: (1) the shape of the word
embedding of a language carries historical and structural 
information, similar to Port et al.  \cite{port2018persistent,port2019topological}'s findings; 
and (2) TDA methods can successfully capture aspects of the shape of language \cite{draganov2024shape}.

Similar to \cite{draganov2024shape,port2018persistent,port2019topological}, 
Dong et al. \cite{dong2024linguistics} extracts the topological
shapes of languages. Specifically, South American languages - 
the Nuclear-Macro-J\^{e} (NMJ) and Quechuan families using TDA.
By using techniques like multiple correspondence analysis (MCA)
for dimension reduction of the categorical-valued dataset and 
persistent homology, Dong et al. \cite{dong2024linguistics} visualizes each language in the 
selected families as a point cloud. This forms the topological shape of the South American languages, such that languages close together are more similar. 
By comparing the topological shapes of the languages, 
it is observed that there are major distinctions between 
the J\^{e}-proper and the non-J\^{e}-proper languages, as well as 
the northern and southern Quechuan languages \cite{dong2024linguistics}. 

Fitz et al. \cite{fitz2022shape} introduces a novel terminology - \textit{word manifold}, 
which is a simplicial complex, whose topological space captures grammatical
structure expressed by the corpus. 
This is done by implementing a technique for generating
topological structure directly from strings of words.
Experiments reveal that the homotopy type of the
word manifold is also influenced by linguistic structure \cite{fitz2022shape}.
Finally, Bouazzaoui et al. \cite{bouazzaoui2021application} explores the topological 
similarity of the shapes of two writing systems - Tifinagh and Phoenician
scripts.

\begin{figure}%[18]{r}{0.50\textwidth}
    \centering
    % \vspace{-25pt}
    \includegraphics[width=0.8\linewidth]{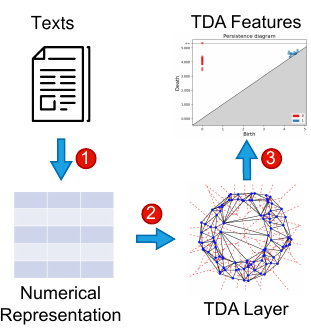}
    \caption{Illustration (inspired by \cite{uchendu2023topformer}) of the Non-theoretical approach of using TDA as a feature extractor in NLP with three steps: 
    (1)-extracting numerical representations, (2)-reformatting for TDA's inputs, and (3)-extracting TDA features.}
    \label{fig:pipeline}
\end{figure}

\begin{figure*}[tb!]
    \centering
    \includegraphics[width=0.8\linewidth]{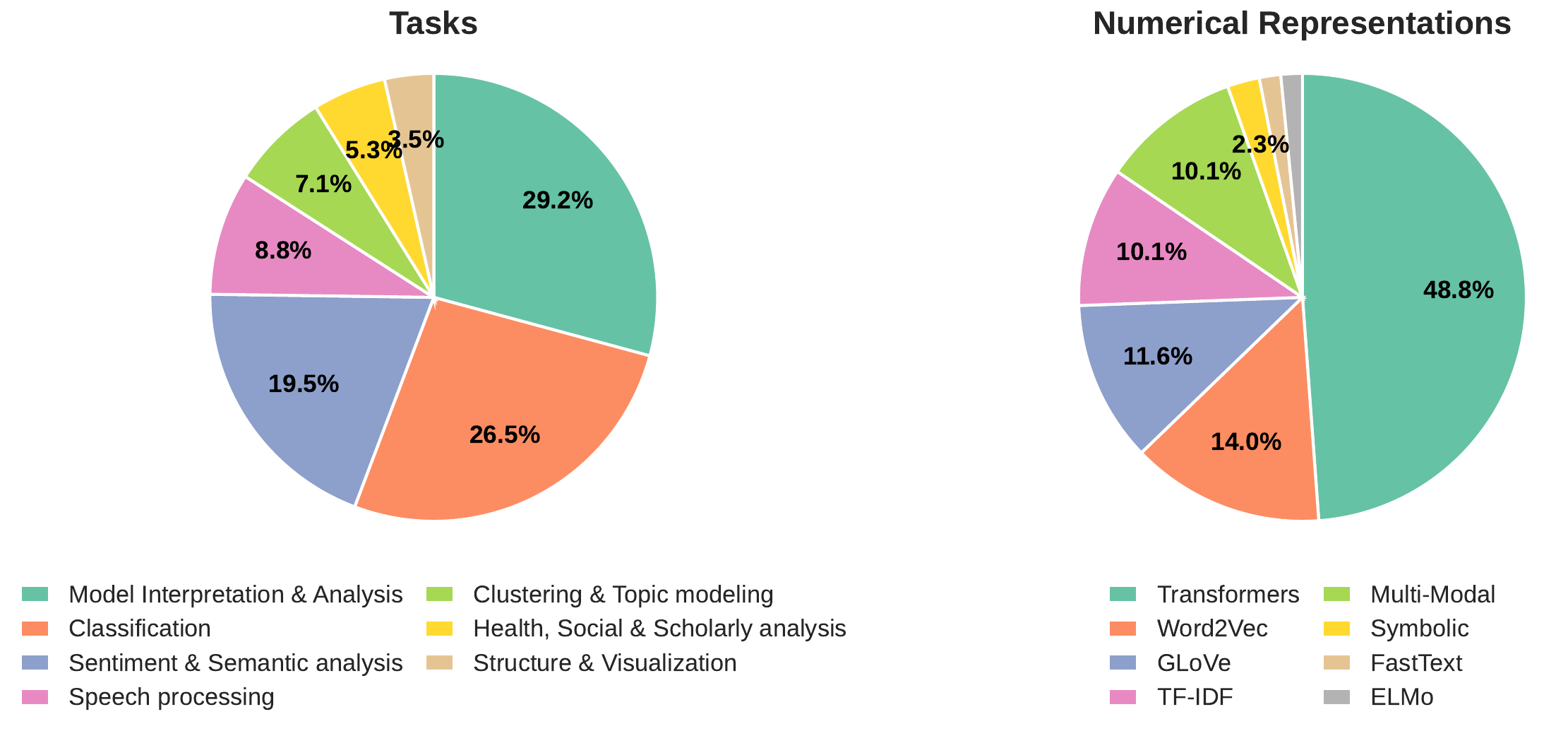}
    \caption{Distribution of the number of publication for each task (\textbf{Left}) and numerical representation type (\textbf{Right}).}
    \label{fig:nontheo_tasks_numeric}
\end{figure*}

% \citet{wei2023topological} employs persistent homology in a spectral geometric problem to answer the question - \textit{Can one hear the shape of a drum?}

\begin{myBox}[]{Definition}%{AA}
\textsc{\textbf{Insight (Shape of Words).}} 
% \footnotesize
\textit{The topological ``shape'' of words is a concept that has interested several linguists, as it can be used to confirm and discover linguistic phenomena within languages. 
It is 
% the most studied theoretical approach, with the main application 
focused on capturing the shape of several languages. 
This concept combines all other frameworks like the semantic and syntactic topological spaces to capture a linguistically informed topological shape of words. 
}
\end{myBox}

% https://umbc-my.sharepoint.com/:p:/g/personal/adaku2_umbc_edu/EXx1o01hthhLuNiIG4c-uLwB3P6BbItMumBE_sSNYMFxvQ?rtime=MaWxbNHb3Eg
% \begin{figure}%[!htb]
%     \centering
%     \includegraphics[width=0.5\linewidth]{figures/TDA_survey_Page4.pdf}
%     \caption{Illustration of the TDA feature extraction pipeline in NLP with three steps: \circled{1}-extracting numerical representations, \circled{2}-reformatting for TDA's inputs, and \circled{3}-extracting TDA features.}
%     \label{fig:pipeline}
%     % \vspace{-10pt}
% \end{figure}

% https://drive.google.com/file/d/1oryy-ORVs0PEVcFYb6wmMYJKSh1fqW42/view?usp=sharing

\section{Non-theoretical Approaches}

There are several ways to categorize the applied/non-theoretical TDA applications in NLP. These applications can be categorized by \textit{task}, 
\textit{learning type}, \textit{data modality}, \textit{TDA technique}, and \textit{numerical representation}. We observe that categorizing these TDA applications by task and 
numerical representation is more meaningful than the other 
categories, since those categories are binary and not 
very descriptive of the landscape. Out of these dimensions, the numerical representation showcases the bottleneck for extracting useful TDA features. See Figure \ref{fig:pipeline} for an illustration of the pipeline for extracting TDA features from numerically represented texts. 
In addition, while we focus on both task and numerical representation, our main taxonomy for the non-theoretical applications is centered on how 
TDA features are extracted from 
different forms of 
numerical representations.
Part of Figure \ref{fig:flowchart} illustrates the taxonomy of non-theoretical 
applications of TDA in NLP tasks. 
See Table \ref{tab:nontheory_label} and \ref{tab:nontheory_labelb} in Appendix for the list of non-theoretical approaches.

% We can categorize non-theoretical TDA applications in NLP by \textit{learning types}, \textit{Modality}, and \textit{TDA techniques}. 
\textit{Learning types} have supervised \cite{elyasi2019introduction,lavery2024combining}, and unsupervised \cite{spannaus2024topological,bonafos2024dirichlet}; 
\textit{Modality} has text \cite{triki2021analysis,kostenok2023uncertainty}, and Speech \cite{ sassone2022bridging}; and 
\textit{TDA techniques}, have Persistent Homology 
\cite{torres2015topic,cherniavskii2022acceptability}, and Mapper \cite{holmes2020topological,elyasi2019introduction}.
% Lastly, see Figure \ref{fig:nontheo_taxo_figs} for the number of publications for Modality, and TDA techniques. 
For data modality, 90\% of applications are concentrated in Text, 
and for TDA techniques 89\% of applications are concentrated in persistent homology. In this survey, we focus on two other broad categories, \textit{Tasks} and \textit{Numerical Representation}, where the connections to prior work are richer.

\subsection{Tasks} \label{sec:tasks}
Tasks in this context are defined as the problem for which a solution is attempted. 
We categorize these NLP problems that TDA practitioners have attempted into seven categories below. We refer the readers to Figure \ref{fig:nontheo_tasks_numeric} (left) for the distribution of the number of publications per task.

\begin{enumerate}[leftmargin=\dimexpr\parindent+0.8\labelwidth\relax,noitemsep,topsep=0pt]
    \item \textbf{Classification}: 
    The most popular application is {deepfake text detection} \cite{lovlie2023text,tulchinskii2024intrinsic,kushnareva2023artificial,kushnareva2021artificial,uchendu2023topformer,wei2025short}. 
    
    \item \textbf{Clustering and Topic Modeling}:
    The most popular application is document clustering and topic modeling
    \cite{holmes2020topological,guan2016topological}.
    
    \item \textbf{Sentiment and Semantic Analysis}: 
    The most popular applications are {linguistic/grammatical acceptability} \cite{cherniavskii2022acceptability,jain2024beyond}, 
    {word sense induction \& disambiguation} \cite{rawson2022topological,temvcinas2018local}, and
    {polysemy word classification} \cite{jakubowskietal2020topology,shehu2024analysis}. 
    
    \item \textbf{Structure and Visualization}:
    The most popular is using Mapper to visualize model hidden weights \cite{garcia2022applications,rathore2023topobert}. 
    
    \item \textbf{Health, Social, and Scholarly Analysis}:
    Since this is not a popular application for TDA, the most interesting applications are - prediction of
    epidemics \cite{lordgrilo} and categorization of lonely people \cite{effahtopological}.
        
    \item \textbf{Speech Processing}: 
    The most popular applications are studying vocalizations \cite{bonafos2023topological,bonafos2024dirichlet}. 
    % music classification~\cite{bergomi2015dynamical,sassone2022bridging}.
        
    \item \textbf{Model Interpretation and Analysis}:
    The most popular applications are model probing to reveal behavior in hidden weights \cite{kostenok2023uncertainty,gourgouliasestimating}.
\end{enumerate}

\subsection{Numerical Representation}
See Figure \ref{fig:nontheo_tasks_numeric} (right) for the distribution of the number of applications for each numerical representation.

\subsubsection{TF-IDF} \label{sec:tfidf}
\textbf{TF-IDF (Term Frequency - Inverse Document Frequency)} is a well-known statistical formula that calculates the importance of words relative to a corpus. 
A few works investigated the extraction of topological features from TF-IDF representations as part of the pipeline illustrated in Figure \ref{fig:pipeline}. 
For instance, SIFT, a persistent \\homology-based model with TF-IDF, is developed to differentiate between child and adolescent writings~\cite{zhu2013persistent}. 
This model represented the TF-IDF features as a time series, and 
then extracted topological features to enhance text classification.
Several other researchers applied this model to other 
 \textit{classification task}, such as deepfake text detection~\cite{lovlie2023text}, 
presidential election speech attribution \cite{huang2022tda}, 
distinguishing between languages by averaging the persistence landscapes \cite{sovdat2016text}, 
age group categorization of lonely people \cite{effahtopological}, and 
movie genre classification~\cite{doshi2018movie,shin2019genre}.
Additionally, Elyasi et al. \cite{elyasi2019introduction} compares the two popular TDA approaches - 
Persistent Homology and Mapper to classify Persian poems.
Also, using Mapper for the \textit{structure \& visualization tasks}, 
Maadarani et al. \cite{maadarani2020shape} 
explains linguistic properties in poetry writing styles, 
and Van et al. \cite{van2020novel} interprets NLP model behavior. 
Lastly, we observe applications in the \textit{clustering \& topic modeling task} - 
keyphrase extraction \cite{guan2016topological}, 
text summarization \cite{kumar2022extractive},
and twitter topic detection \cite{torres2015topic}; 
\textit{sentiment \& semantic analysis task} - 
legal entailment \cite{savle2019topological}, and sentiment analysis of movie reviews \cite{michel2017does}.

% \begin{myBox}[]{Definition}%{AA}
% \textsc{\textbf{Insight (TFIDF).}} 
% % \footnotesize
% \textit{
% Topological features extracted from TF-IDF have been
% applied to four out of the seven tasks - classification, clustering \& topic modeling, 
% sentiment \& semantic analysis, and structure \& visualization. 
% This suggests that while there has been an exponential improvement in contextual and non-contextual word embeddings for numerically representing texts, TF-IDF can still extract decent features from text that can be further leveraged by TDA.}
% \end{myBox}

\subsubsection{Pre-trained Non-contextual Embeddings}

\noindent \textbf{Word2Vec Embeddings.} Word2Vec embeddings are a type of word representation that allows words with similar meanings to have similar vector representations~\cite{mikolov2013distributed}. 
Thus, we observe applications in the \textit{structure \& visualization task}, where Haghighatkhah et al. \cite{haghighatkhah2022story} 
creates story trees to trace story lines.
Next, we observe applications in \textit{sentiment \& semantic analysis task}, where TDA is applied to novel problems such as 
the creation of a topologically-enhanced search engine using Mapper \cite{Filco306}, 
measuring distance between the literary style of Spanish poets - Francisco de Quevedo, Luis de G\'{o}ngora, and Lope de Vega \cite{paluzo2019towards}, 
detecting narrative shifts in media discourse \cite{bailey2025detecting}, 
distinguishing news articles and poems by detecting plot holes \cite{alimpiev2020plot}, 
analysis of contradictions within texts \cite{wu2022topological}, 
and
word sense induction and disambiguation~\cite{rawson2022topological,temvcinas2018local}.
For the \textit{classification} task, researchers detect fraudulent papers~\cite{tymochko2021connections}, and
topological loops in logical statements~\cite{tymochko2020argumentative}. 

Furthermore, we observe applications in the 
\textit{health, social, and scholarly analysis} task - 
disease prediction from epidemic \\curves~\cite{lordgrilo}, and 
Yadav et al. \cite{yadav2025missing} uses doc2vec to create top2vec (i.e., topological features $\to$ vec) of publication documents to use the holes found with persistent homology to determine missing documents vs. innovative research.
Additionally, \cite{holmes2020topological,wright2020topological} perform \textit{topic modeling tasks}. Finally, for the \textit{model interpretation and analysis} task, 
Feng et al. \cite{feng2024geometry} uses both topological and geometrical features to investigate the quality of LLM-enhanced 
data augmentation, 
Sun et al. \cite{sun2023topological}
derives the correlation between sentence vectors and their semantics, and 
Yessenbayev et al. \cite{yessenbayev2022comparison,yessenbayev2024use}
compares the semantic alignment of text and speech embeddings for text-speech pairs. 

% \begin{myBox}[]{Definition}%{AA}
% \textsc{\textbf{Insight (Word2Vec).}} 
% % \footnotesize
% \textit{Topological features extracted from Word2vec are considered rich, such that researchers have attempted six out of seven tasks. 
% The only task that has not been attempted is speech processing, since a different embedding is needed for audio data. 
% This suggests that word2vec embeddings capture enough linguistic features that, when augmented with topological features 
% can improve baseline performance.
% Finally, we observe that the sentiment and semantic analysis task is the most popular task for word2vec. 
% }
% \end{myBox}

% \vspace{20pt}
\noindent \textbf{GloVe Embeddings.} 
GloVe or Global Vectors for Word Representation is another technique for numerically representing texts as embeddings. 
Topological features extracted from GloVe embeddings have been applied to the following tasks - 
\textit{classification task}, which include 
author attribution of novelists \cite{gholizadeh2018topological}, 
fake news detection \cite{deng2022topological}, and 
deepfake text detection~\cite{lovlie2023text,alanis2025detection}; 
\textit{sentiment \& semantic analysis task}, which include 
document categorization \cite{gholizadeh2020novel}, and
capturing circles in circular arguments \cite{tymochko2020argumentative,zadrozny2021note};
\textit{model interpretation and analysis}, 
where both Haim et al. \cite{haimmeirombobrowski2022unsupervised} and Michel et al. \cite{michel2017does}
compare text representations \& embeddings, Spannaus et al.  \cite{spannaus2024topological} explains model performance, and
Zadrozny et al. \cite{zadrozny2021abstraction} tests the manifestation of intelligence and understanding in models; and
\textit{health, social, and scholarly analysis task}, which include 
social anxiety detection \cite{byers2021hidden}, 
and keywords extraction of scholarly documents \cite{novak2019network}.

% \begin{myBox}[]{Definition}%{AA}
% \textsc{\textbf{Insight (GloVe).}} 
% % \footnotesize
% \textit{Topological features extracted from GloVe are considered rich, but not as rich as Word2Vec since only four out of seven tasks are attempted. Tasks, not attempted are:  (1) speech processing, (2) structure \& visualization, and (3) clustering \& topic modeling. 
% The most popular application is in the model interpretation \& analysis task, suggesting that glove embeddings can be probed using TDA techniques to excavate interpretations of model performance. 
% }
% \end{myBox}
% \vspace{5pt}

\noindent \textbf{FastText Embeddings.}
FastText embeddings are built on the \\Word2Vec approach by incorporating subword information, improving the representation of rare words, and allowing for embedding out-of-vocabulary words \cite{bojanowski2017enriching}. 
This type of embedding is not a popular feature extractor that researchers employ to enhance topological features, as only two tasks are attempted - 
\textit{sentiment \& semantic analysis task}, 
which include polysemy word classification \cite{jakubowskietal2020topology,triki2021analysis,shehu2024analysis}, and 
word sense induction \& disambiguation \cite{jakubowskietal2020topology}; 
and \textit{text classification}, where Tymochko et al. 
\cite{tymochko2021connections} detects fraudulent papers. 

% \begin{myBox}[]{Definition}%{AA}
% \textsc{\textbf{Insight (FastText).}} 
% % \footnotesize
% \textit{FastText is not widely adopted for topological applications, with only a few researchers 
% applying it to two tasks - text classification and sentiment \& semantic analysis. 
% Sentiment \& Semantic analysis is the more popular application involving polysemy word classification and word sense induction \& disambiguation. 
% % This could be because FastText is not widely adopted by the NLP community. 
% }
% \end{myBox}

\subsubsection{Pre-trained Contextual Embeddings}

\noindent \textbf{Transformer Embeddings.} 
Researchers have evaluated the \\strength of the TDA features extracted from Transformer-based \cite{vaswani2017attention} embeddings. 
Using the idea of self-attention, the neural network can encode more semantic and syntactic features than previous embeddings, which should allow for 
richer TDA features to be extracted. 
%Typically, for classification problems the output from the last hidden layer, known as the $pooled\_output$ is the input to the softmax layer.
To incorporate TDA features for various tasks, several researchers have investigated the efficacy of using other outputs of encoder and decoder Transformer models - \textit{CLS Embedding output}, \textit{hidden weights}, and \textit{attention weights} to extract high-quality additional features.

\noindent \textbf{\texttt{CLS Embedding Output.}}
Researchers have applied these features on 
\textit{text classification task}, specifically for deepfake text detection \cite{tulchinskii2024intrinsic,kushnareva2023artificial,wei2025short,guilinger2025topological}, 
fake news detection~\cite{lavery2024combining}, and
TEDtalk public speaking ratings classification~\cite{das2021persistence}. 
Additionally, we observe applications to the \textit{topic modeling task}~\cite{byrne2022topic,hopp2025persistent}. 
Next, for the \textit{model interpretation \& analysis task}, Gourgoulia et al. \cite{gourgouliasestimating} probes LLMs to estimate class separability of text datasets, and 
Proskura et al. \cite{proskura2024beyond} uses topological information from encoder models to select the best models to use when building an ensemble.
In addition, 
Rair et al. \cite{rair2025annotators} uses Mapper to visualize 
how fine-tuning of models like RoBERTa-Large restructures the 
embedding space into modular, non-convex regions to align with model predictions. This technique aims to 
visualize the geometry and topology of when annotators agree and disagree \cite{rair2025annotators}.
Furthermore, Arun et al. \cite{arun2025topo} employs TDA for the \textit{structure \& semantic task}, 
detecting controversial vs. non-controversial political discourse by capturing the shifts in discourse for controversial data. 
Similarly, Meng et al. \cite{meng2024encoding}
performed a \textit{semantic task} of using persistent homology to augment and improve personalized web search.
Next, Chandra et al. \cite{chandra2026topology} performs a \textit{health analysis} task using persistent homology to 
track mental health journeys in online communities.
Finally, Rathore et al. \cite{rathore2023topobert} performs the \textit{structure \& visualization task} in combination with the \textit{model interpretation \& analysis task} by visualizing the 
training process of transformer-based models.

% \begin{myBox}[hb]{Definition}%{AA}
% \textsc{\textbf{Insight (Transformer-CLS).}} 
% % \footnotesize
% \textit{CLS embedding is applied to six tasks - 
% text classification (the most popular), 
% topic modeling, model interpretation \& analysis,
% sentiment \& semantic analysis, 
% health analysis, and 
% structure \& visualization. 
% The most interesting subtasks are probing LLMs to estimate class separability of text datasets and visualizing the training process of transformer-based models. 
% }
% \end{myBox}

% \citet{lavery2024combining} performs fake news detection with a topologically-augmented BERT model. 
% \citet{proskura2024beyond} selects the best models to create 
% an ensemble model with by first checking the topological similarity of the embeddings of the 
% models. 
% \citet{das2021persistence} extracts topological features using the sentence embeddings of BERT to represent texts as persistent image vectors to perform TEDtalk public speaking ratings classification. 
% \citet{gourgouliasestimating} proposes an unsupervised approach for estimating the class separability of text datasets for binary and multi-class tasks. 
% \citet{rathore2023topobert} proposes \textit{TopoBERT}, a visual tool that topologically analyzes 
% the fine-tuning process of the Transformer-based model using Mapper. 
% Similarly, \citet{byrne2022topic} uses Mapper for unsupervised 
% topic modeling by representing the embeddings as a graph. 

% \subparagraph{Pooled\_output}
\noindent \textbf{\texttt{Hidden Weights.}}
\textit{Classification task} include deepfake text detection \cite{uchendu2023topformer,rejimoan2025detection}, 
language translation \cite{asriani2025nmt},
and 
code attribution \cite{macdonald2024binary}.
Next, Garcia et al. \cite{garcia2022applications} explores a combination of the \textit{sentiment \& semantic analysis} and \textit{structure \& visualization} tasks by using Mapper to visualize polysemous words in 
the hidden representations of the BERT transformer model.
Bensalem et al. \cite{bensalem2025detecting}
performs a \textit{sentiment analysis} task by 
using the sentiment scores of original vs. translated texts extracted from a Transformer-based model.
They represent these scores in a time series form and use 
zigzag persistent homology to detect 
sentiment shift in translated texts. 
Similarly, Goshev et al. \cite{goshevtopology} investigates the semantic topology of sentences encoded by transformer embeddings. 
% They explore whether persistent homology of 
% the embeddings capture the geometry of the
% embedding space \cite{goshevtopology}.
Zhang et al. \cite{zhang2026text} uses persistent homology to improve text summarization by capturing the global structure of texts.
Alexander et al. \cite{alexander2023topological} combines the \textit{health analysis} and \textit{visualization} tasks
to visualize GPT-3's embeddings of hate speech, misinformation, 
and psychiatric disorder texts with Mapper.  
Ruppik et al. \cite{ruppik2024local} performs the \textit{clustering \& topic modeling task} through dialogue term extraction. 

The rest of the applications attempt the 
\textit{model interpretation \& analysis task}, making it the most popular task.
Gardinazzi et al. \cite{gardinazzi2024persistent} proposes a novel metric - \textit{persistence similarity} to prune redundant layers in LLMs.
Zheng et al. \cite{zheng2026topology} uses the same technique to prune large vision-language models. 
Balderas et al. \cite{balderas2025green} proposes Persistent BERT Compression and Explainability (PBCE) to compress BERT by pruning redundant layers.
Sun et al. \cite{sun2023topological} probes the correlation between sentence vectors and their semantics.
Garcia et al. \cite{garcia2024relative} performs zero-shot model stitching by employing \textit{topological densification} (i.e., creating a topology-aware loss function).
Huang et al. \cite{topologyloss2025} uses a topology-aware loss function to improve prompt tuning.
Athreya et al. \cite{athreya2025hole} proposes a framework - HOLE (Homological Observation of Latent Embeddings for Neural Network Interpretability) to visualize the latent space of BERT for Named Entity Recognition (NER) task.

Next, several researchers probe the reasoning processes of LLMs: 
(1) Tan et al. \cite{tan2025shape} captures the geometry and topology of reasoning in LLMs using a mathematics examination dataset to capture step-by-step reasoning for solving non-trivial word problems;
(2) Li et al. \cite{li2025understanding} and 
Zhang et al. \cite{zhang2026tda} probe the reasoning process of the latent space of LLMs when using the chain-of-thought (CoT), tree-of-thought, and graph-of-thought prompts;
(3) More et al. \cite{more2025optimizing} proposes Enhanced Dirichlet
and Topology Risk (EDTR), which is a 
novel decoding strategy that leverages persistent homology and 
Dirichlet-based uncertainty quantification to calculate 
LLM confidence;
(4) Zhang et al. \cite{zhang2026learning} proposes GHS-TDA, a reasoning technique to improve CoT reasoning by constructing 
a semantically enriched global hypothesis graph using persistent homology to capture global structures and remove redundancies; and 
(5) Ishimtsev et al. \cite{ishimtsevhole} proposes a theory that consciousness might appear when a thinking process loops back on itself with CoT prompting of LLMs. 
Using this theory, Ishimtsev et al. \cite{ishimtsevhole} implements a topology analogy, guided by persistent homology to define: 
normal reasoning as a \textit{straight path},
self-reflection as a \textit{loop}, 
and 
consciousness as a \textit{stable loop around a hole} \cite{ishimtsevhole}.

Finally, in support of the \textit{model interpretation \& analysis task}, several researchers probe the embedding space of LLMs to quantify the topology of the latent space. 
Fitz et al. \cite{fitz2024hidden} measures the topological complexity (known as \textit{perforation}) of the hidden representation of LLMs to understand their topological shapes.
Also, Fitz et al. \cite{fitz2023large} investigates the topological structure of the brain of ChatGPT concerning its notion of fairness.
Chauhan et al. \cite{chauhan2022bertops} proposes a novel scoring metric - 
\textit{persistence scoring function} which captures the homology of the hidden representations of BERT.
Fay et al. \cite{fay2025holes} investigates the differences in the topological structure of the latent space of adversarial vs. non-adversarial texts in LLMs.
Kudriashov et al.  
\cite{kudriashov2024more} probes BERT's hidden weights 
on new grammatical features, known as \textit{polypersonality}. 
Lastly, Yan et al. \cite{yan2025explainablemapperchartingllm} builds an \textit{Explainable Mapper} framework that uses two mapper agents to probe the embedding space of language models, 
and generate readable linguistic explanations using summarization, comparison, and perturbation operations.

\begin{figure*}[!htb]
    \centering
    \includegraphics[width=1\linewidth]{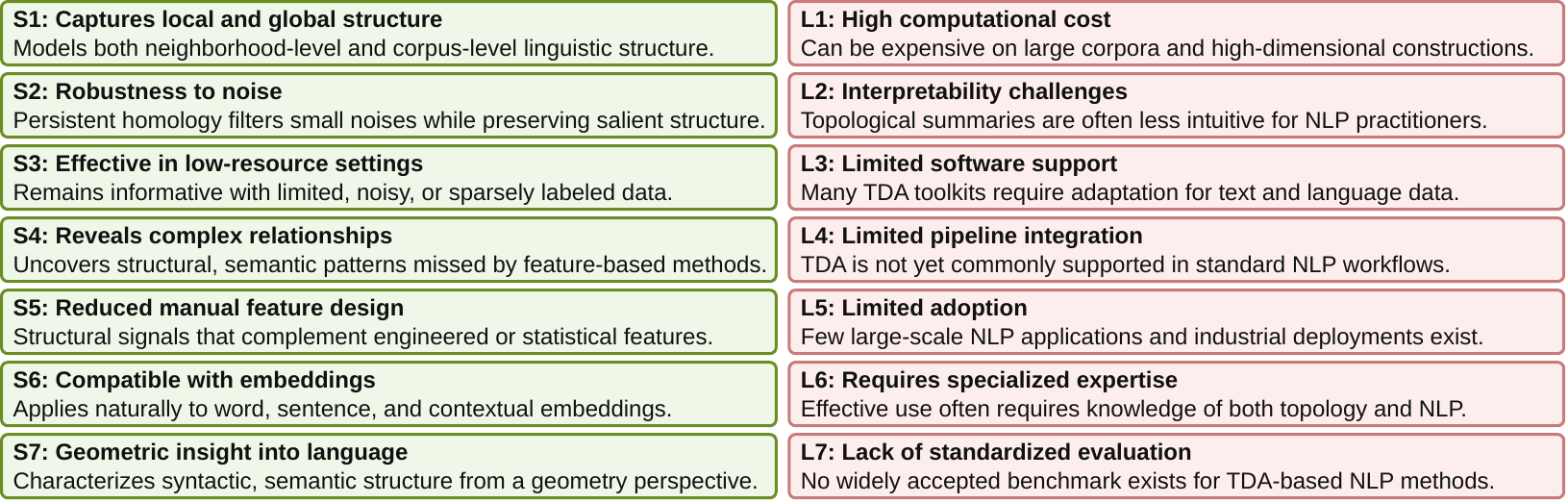}
    \caption{Strengths (\textbf{S}) and limitations (\textbf{L}) of applying TDA in NLP.}
    \label{fig:strength_limit}
    % \vspace{-10pt}
\end{figure*}

\begin{table*}[!htb]
\centering
\small
\setlength{\tabcolsep}{8pt}
\renewcommand{\arraystretch}{1.15}
\begin{threeparttable}
\begin{tabularx}{\textwidth}{
>{\raggedright\arraybackslash}p{0.07\textwidth}
>{\raggedright\arraybackslash}p{0.24\textwidth}
>{\raggedright\arraybackslash}X}
    \toprule
    \textbf{Category*} & \textbf{Method} & \textbf{Description} \\
    \midrule
    
    Linear
    & PCA
    & Reduces dimensionality while preserving variance through orthogonal transformations. \\
    
    Linear
    & MDS
    & Projects high-dimensional data into lower dimensions by preserving pairwise distances. \\
    
    Nonlinear
    & t-SNE
    & Projects high-dimensional data into 2D or 3D while preserving local relationships. \\
    
    Nonlinear
    & UMAP
    & Similar to t-SNE, but faster and often better at preserving global structure. \\
    
    Nonlinear
    & Isomap, LLE
    & Captures intrinsic structure in high-dimensional data using graph-based techniques. \\
    
    Neural
    & Autoencoders
    & Learn compressed data representations through encoding and decoding. \\
    
    Neural
    & Geometric Deep Learning
    & Applies neural networks to non-Euclidean spaces such as graphs and manifolds. \\
    
    Graph
    & Spectral Clustering, GNNs
    & Uses graphs to model relationships and structure within data. \\
    
    Clustering
    & DBSCAN, K-Means, HDBSCAN
    & Groups similar data points based on distance or density. \\
    
    Kernel
    & SVM, Kernel PCA
    & Uses non-linear mappings to extract complex structures in data. \\
    
    Geometric
    & Delaunay Triangulation, Convex Hull
    & Uses geometric techniques to extract data characteristics by analyzing spatial boundaries. \\
    
    \bottomrule
\end{tabularx}
\begin{tablenotes}[flushleft]
\footnotesize
\item \textit{(*) }
\textbf{Linear}: Linear Projection; 
\textbf{Nonlinear}: Nonlinear Projection; 
\textbf{Neural}: Neural Network-based Methods; 
\textbf{Graph}: Graph-based Methods; 
\textbf{Clustering}: Clustering Methods; 
\textbf{Kernel}: Kernel Methods; 
\textbf{Geometric}: Geometric Methods.
\end{tablenotes}
\end{threeparttable}
\caption{Alternatives to Topological Data Analysis in NLP}
\label{tab:tda_alternatives}
\end{table*}

\noindent \textbf{\texttt{Attention Weights.}}
Attention weights extracted from BERT and its variants (e.g., RoBERTa) 
have been transformed to both directed and undirected graphs, on top of which different TDA features are extracted for 
the \textit{text classification task} such as deepfake text detection~\cite{kushnareva2021artificial}, 
LLM hallucination detection \cite{bazarova2025hallucination}, 
Code-LLM hallucination detection \cite{voronkova2026topology},
robustness evaluation of TDA features~\cite{perez2022topological},
authorship attribution of Japanese texts~\cite{sakurai2025authorship},
out-of-distribution detection (OOD)~\cite{pollano2023detecting,perez2022topological}, and
vulnerability detection in code \cite{snopov2024vulnerability}.
Additionally, this framework is applied to the \textit{sentiment \& semantic analysis task}, 
specifically on human linguistic competence (i.e., grammatical acceptability judgment) ~\cite{cherniavskii2022acceptability,proskurina2023can,perez2022topological},
dialog term extraction~\cite{vukovic2022dialogue}, 
and document coherence~\cite{jain2024beyond}.
Similarly, with the same framework, we observe a \textit{speech processing} application~\cite{tulchinskii2023topological}. 
Finally, researchers attempt the \textit{model interpretation \& analysis task}, 
where Kostenok et al. \cite{kostenok2023uncertainty} estimates uncertainty in encoder models; 
Samaga et al. \cite{samaga2026halluzig} characterizes the occurrence of hallucination in LLMs;
Tsai et al. \cite{tsai2026shape} performs the same experiments to investigate the effects of sandbagging and code-injections in language model latent space;
Varadarajan et al. \cite{varadarajan2025augmentingbiasdetectionllms} examines why GPT-2 significantly misrepresents gender and race identity categories by identifying which attention heads 
are responsible for the misclassification of specific identity groups; 
and Proskura et al. \cite{proskura2024beyond} performs dynamic weighting for building ensemble models.

% \begin{myBox}[]{Definition}%{AA}
% \textsc{\textbf{Insight (Transformer-Attention).}} 
% % \footnotesize
% \textit{Given the wealth of information contained in the attention weights, researchers have applied topological features extracted from these weights to several tasks. The most interesting applications are observed on the sentiment and semantic tasks - grammatical acceptability judgment and document coherence analysis. 
% Finally, we observe that for all the Transformer-based weights, the most explored subtask is \underline{deepfake text detection}. 
% }
% \end{myBox}

\noindent \textbf{ELMo Embeddings.}
ELMo embeddings are a type of word representation that captures both the meaning of words and their usage in context \cite{peters2018deep}. 
Similar to other embeddings, ELMo has also been leveraged to extract topological features. 
%We will discuss its usage in TDA in this section. 
Tymochko et al. \cite{tymochko2021connections} performs \textit{text classification} to
detect fraudulent papers by examining their titles and abstracts. 
This is done in comparison of other embeddings (Word2Vec, GloVe, FastText, and Frequency Time Series) to determine the best embeddings to extract strong topological features. Similarly, Alimpiev et al. \cite{alimpiev2020plot}
compares using ELMo, GloVe, Word2Vec, and BERT embeddings to 
perform semantic analysis on news articles and poems.
%in order to categorize sub-groups within them. 

% \begin{myBox}[]{Definition}%{AA}
% \textsc{\textbf{Insight (ELMo).}} 
% % \footnotesize
% \textit{ELMo embeddings is applied to two tasks - text classification and semantic analysis in comparison to other embeddings to determine the best numerical representation for extracting and capturing rich TDA features. 
% }
% \end{myBox}

% \paragraph{LSTM attention Embeddings}
% LSTM Attention Embeddings refer to word or sequence representations generated by Long Short-Term Memory (LSTM) networks integrated with an attention mechanism. This combination leverages the strengths of LSTM in handling sequential data and the attention mechanism's ability to focus on relevant parts of the input sequence. 
% \citet{zadrozny2021abstraction} finds that we need better mathematical models of
% abstraction for examining the ability of neural networks. 
% They investigate the ability of LSTM attention to exhibit System 2 reasoning 
% in well-known manifestations of intelligence and understanding. 
% Using the example of
% the “Look and Say” puzzle, \citet{zadrozny2021abstraction} topologically examines 
% the output when they reverse the puzzle. 
% Findings reveal that neural networks, while their performance is highly accurate, 
% lack comprehension, especially when evaluated on the reverse puzzle. 

\subsubsection{Symbolic Representations}
Symbolic representations in the context of AI and cognitive science refer to the use of symbols such as letters, numbers, tokens, or abstract entities to represent concepts, objects, relationships, and rules within a system. These symbols can be manipulated according to predefined rules to perform reasoning, problem-solving, and decision-making. 
Symbolic representation contrasts with \\sub-symbolic representations, such as neural network-based embeddings, which do not explicitly use symbols or rules.
This section then discusses the creation of symbolic representations by using \textit{principles of letter coding (PLC)}, \textit{principles of speech sound coding (PSSC)}, and one-hot encoding, of which topological features are then extracted. 

PLC refers to rules and methods used to encode letters that fuel various communication systems, cryptography techniques, or linguistic analyses. Letter coding transforms letters or characters into different symbols, numbers, or other forms. PSSC is similar to PLC but for extracting topological features from speech sounds.
One particular application of PLC and PSSC is the study of Ukrainian tongue twisters \cite{yurchuk2023tongue,kovaliuk2024topological}. 
These applications attempt two tasks, \textit{sentiment \& semantic analysis} and \textit{speech processing}, respectively. 
Yurchuk et al. \cite{yurchuk2023tongue} uses the PLC to create word embeddings for Ukrainian tongue twisters and extract topological features from such embeddings with persistent homology. 
This is to distinguish tongue twister from a simple narrative sentence using support vector machine and decision tree classifiers. Similarly, 
Kovaliuk et al. 
\cite{kovaliuk2024topological} uses PSSC for classifying spoken Ukrainian tongue twisters. 
Additionally, Escolar et al. \cite{escolar2025topological} uses one-hot encoding to represent cooking recipes numerically. Using these symbolic representations of recipes, they use the persistent homology's concept of holes to create new recipes, which were implemented and confirmed to be acceptable by a sensory evaluation study \cite{escolar2025topological}.

% \begin{myBox}[]{Definition}%{AA}
% \textsc{\textbf{Insight (Symbolic).}} 
% % \footnotesize
% \textit{Symbolic representations are a novel technique for numerically representing text. The most popular applications of this technique are on Uranian tongue twisters, both for the sentiment \& semantic analysis and speech processing tasks. 
% This suggests that such a technique could be applied to processing 
% low-resource languages, where the current popular numerical representation may be insufficient. 
% }
% \end{myBox}

\subsubsection{Multi-Modal Representations}
TDA features have also been extracted from other representations of NLP-related features, including multimedia data, such as audio and video. 
In this section, the most popular task performed by researchers is 
\textit{speech processing}, where applications include - 
studying human vowels and infant vocalizations \cite{bonafos2023topological,bonafos2024dirichlet}, 
speech recognition \cite{yu2025topologicaldeeplearningspeech, lameris2024topological}, 
emotion recognition from audio speech~\cite{gonzalez2019towards,raskin2025shape} and audio in videos~\cite{paluzo2022emotion}, 
depression detection from audio clips~\cite{tlachac2020topological},
recognizing voiced and voiceless consonants in \\speech~\cite{zhu2024topology}.
% music classification~\cite{bergomi2015dynamical,sassone2022bridging},

Next, we observe several applications with Vision-Language Models (VLMs), which also attempt the \textit{model interpretation \& analysis task} - 
integrating Representation Topology Divergence (RTD) with the loss function to align the topological structures of image and text representations during tuning \cite{huang2025topology}; 
implementing a topological approach to align the image and text latent manifolds in VLMs \cite{zhang2024homology,rahim2024topological,xu2026topo}; 
assessing the adversarial robustness of VLMs by measuring topological consistency \cite{vu2025topological};
assessing the topological alignment in VLMs
between images and multi-lingual text (i.e., French, Spanish, Russian, etc.) \cite{youtopological}; and 
multimodal recommendations improvements
\cite{bachiri2025topological}. 

\definecolor{rootblue}{HTML}{2F5AA8}
\definecolor{rootbluefill}{HTML}{EAF1FF}

\definecolor{foundfill}{HTML}{E8F0FE}
\definecolor{founddraw}{HTML}{5B8DEF}

\definecolor{methfill}{HTML}{F3E8FF}
\definecolor{methdraw}{HTML}{9B6DDB}

\definecolor{frontfill}{HTML}{E6F7F5}
\definecolor{frontdraw}{HTML}{2BAE9A}

\definecolor{softgray}{HTML}{6B7280}

\begin{figure*}[!htb]
\centering
\resizebox{0.98\textwidth}{!}{%
\begin{tikzpicture}[
    font=\large,
    >=Latex,
    rootbox/.style={
        draw=rootblue,
        fill=rootbluefill,
        rounded corners=6pt,
        line width=1.4pt,
        minimum width=12.6cm,
        minimum height=1.45cm,
        align=center,
        font=\bfseries\LARGE,
        text=rootblue
    },
    foundbox/.style={
        draw=founddraw,
        fill=foundfill,
        rounded corners=5pt,
        line width=1.2pt,
        minimum width=5.6cm,
        minimum height=1.3cm,
        align=center,
        font=\bfseries\Large,
        text=black
    },
    methbox/.style={
        draw=methdraw,
        fill=methfill,
        rounded corners=5pt,
        line width=1.2pt,
        minimum width=5.9cm,
        minimum height=1.3cm,
        align=center,
        font=\bfseries\Large,
        text=black
    },
    frontbox/.style={
        draw=frontdraw,
        fill=frontfill,
        rounded corners=5pt,
        line width=1.2pt,
        minimum width=5.6cm,
        minimum height=1.3cm,
        align=center,
        font=\bfseries\Large,
        text=black
    }
]

% Root
\node[rootbox] (root) {Open Problems in Topological Data Analysis for NLP};

% Category boxes
\node[foundbox, below left=1.8cm and 5.7cm of root] (found) {Foundational Enablers};
\node[methbox, below=1.8cm of root] (meth) {Methodological Challenges};
\node[frontbox, below right=1.8cm and 5.7cm of root] (front) {Frontier Expansions};

% Main connections
\draw[->, line width=1.15pt, draw=rootblue] 
(root.south west) .. controls +(down:0.9) and +(up:0.9) .. (found.north);

\draw[->, line width=1.15pt, draw=rootblue] 
(root.south) -- (meth.north);

\draw[->, line width=1.15pt, draw=rootblue] 
(root.south east) .. controls +(down:0.9) and +(up:0.9) .. (front.north);

% Cross-category relations
\draw[->, dashed, line width=1.0pt, draw=softgray] 
(found.east) -- node[above, font=\bfseries\LARGE, text=softgray] {supports} (meth.west);

\draw[->, dashed, line width=1.0pt, draw=softgray] 
(meth.east) -- node[above, font=\bfseries\LARGE, text=softgray] {enables} (front.west);

\end{tikzpicture}%
}
\caption{A high-level taxonomy of open problems in applying topological data analysis to NLP, organized into foundational enablers, methodological challenges, and frontier expansions.}
\label{fig:open_problems_taxonomy}
\end{figure*}

\section{Benefits and Challenges of TDA}\label{sec:tda_others}
The use of TDA in NLP presents both clear opportunities and important challenges, many of which also extend to adjacent topological approaches. Figure \ref{fig:strength_limit} summarizes the main strengths and limitations of applying TDA across NLP tasks.
Among its most salient strengths is its effectiveness in low-resource settings, which makes it particularly attractive for noisy, limited, and heterogeneous data. 
%More broadly, TDA provides a principled way to capture local and global textual structure, uncover complex relationships, and maintain robustness to noise. 
Although TDA has been shown to be a powerful mathematical technique by the numerous applications discussed above, 
there are alternatives to TDA (Table \ref{tab:tda_alternatives}).
While these alternatives have shown great utility in a variety of tasks, such as dimension reduction \cite{binnie2025surveydimensionestimationmethods},
TDA is the only technique that can extract not only local but global features.
More broadly, TDA provides a principled way of capturing local and global textual structure, uncovering complex relationships, and maintaining robustness to noise.
At the same time, its wider adoption remains constrained by several practical barriers, including high computational cost, limited interpretability, relatively immature software support, and the need for specialized expertise.

However, while persistence diagrams are not intuitively interpretable, 
in the context of NLP,
0D features (connected components) corresponds to clusters of semantically similar documents (topics), while 1D features (loops) can capture transitional or overlapping themes where documents form continuums rather than discrete groups. These structures help identify stable topic groupings, semantic relationships, and potential outliers, with robustness to noise.
Compared to PCA or t-SNE, persistence diagrams provide complementary insights. PCA captures linear variance and may miss nonlinear structure, while t-SNE emphasizes local clustering but can distort global relationships. In contrast, persistence diagrams analyze the data in its original space and quantify both local and global structure across scales.

These strengths are further reflected in a broader body of work on related topological approaches in NLP. Beyond Persistent Homology and Mapper, researchers have explored methods grounded in simplicial homology, morse theory, homotopy, and other notions of connectedness and shape to extract structural information from language data. Although these methods are often less formalized than Persistent Homology, they similarly demonstrate the value of topological thinking for modeling semantic structure, discourse organization, stylistic variation, and model behavior. For example, topological notions such as the connected component dimension ($\beta_0$) have been used to assess document coherence through semantic connectedness \cite{chiang2007discover}, while more recent work has extended these ideas to dialogue semantics \cite{santacana2025topological} and to topological semantic spaces for studying sociolinguistic phenomena in LLMs \cite{ionescu2025generative}. Other studies treat text as an object with shape, using topology to characterize stylistic and linguistic patterns \cite{luong2007trees} or to quantify semantic differences between real and fake news through thematic complexity and connectedness \cite{scalvini2025semantic}. More recently, topological ideas have also appeared in safety-oriented LLM research, where homotopy-based methods are used to obfuscate malicious prompts or optimize jailbreak attacks \cite{lazo2026llm,wangfunctional2025}.

Taken together, these studies suggest that the appeal of TDA in NLP is part of a broader methodological advantage of topological approaches: their ability to reveal structural regularities in language that are often difficult to capture through conventional feature-based or purely predictive methods alone. 
Similarly, limitations of TDA, such as computational expense, interpretability challenges, and limited software, highlight the practical obstacles that must be addressed before topological methods can see broader adoption in NLP. While ongoing advances in algorithms, software, and computing infrastructure offer reasons for optimism, these strengths and challenges together point to several promising directions for future work, which we discuss in Section \ref{sec:future}.

\section{Open Problems}\label{sec:future}
% We discuss the open problems and future directions for TDA applications in NLP, 
% as well as ways in which researchers can leverage these benefits and mitigate these risks. 
Although TDA has shown growing promise in NLP, many challenges remain before its full potential can be realized. In this section, we organize the main open problems and future directions into three broad categories: foundational enablers, methodological challenges, and frontier expansions. Together, these categories highlight how researchers can better leverage the strengths of TDA while addressing its current limitations and risks (Figure \ref{fig:open_problems_taxonomy}).

\subsection{Foundational Enablers}

\noindent \textbf{LLM-Assisted TDA Code Generation.}
One of the major challenges in applying TDA to NLP tasks is the steep learning curve associated with its mathematical foundations, which are often accessible only to expert audiences. Moreover, theorists who develop and understand these advanced concepts do not always collaborate with computational scientists to translate them into executable code. 
To address this gap, Liu et al. \cite{liu2024chatgpt} proposes using ChatGPT to generate Python code for TDA concepts by training it on these mathematical foundations. Their findings suggest that ChatGPT can alleviate this bottleneck, particularly for complex TDA concepts like hypergraphs, digraphs, and persistent harmonic space, which have not been as heavily explored as the Vietoris-Rips complex \cite{liu2024chatgpt}.
Similarly, experts can develop specialized code generators, such as fine-tuning models like Code-Llama \cite{roziere2023code} on TDA concepts. 
In addition, Li et al. \cite{li2026large} proposes to benchmark LLM as topological thinkers by giving them tasks to implement persistent homology on. 
The idea is to create \textit{LLM4PH}, an LLM for persistent homology. 
Therefore, we observe that we are closer to 
creating a dedicated LLM for TDA code generation 
which could significantly lower the barrier to entry, encouraging the NLP community to explore TDA applications more innovatively.

\noindent \textbf{Topology-Linguistics Alignment.}
TDA can come across as not intuitive. Even though it has the potential to be applied to interpret different behaviors of modern NLP models, there is still a need for theoretical approaches that better tie TDA features to linguistic phenomena 
but intuitively. 
% There is a need for theoretical approaches that better tie TDA features to linguistic phenomena. 
%, mechanistic explanation on how that process happens. 
% \cite{rathore2023topobert} tasks, TDA being applied to such tasks remains an open problem. This is even more glaring in the NLP domain, where texts take on the shape of their numerical representation. Thus, we conclude that to successfully implement the use of TDA in explaining or interpreting model performance on NLP tasks, we need more theoretical applications. This is because the theoretical applications can better tie TDA features to linguistic phenomena, which is necessary to achieve explainability. 
For example, Draganov et al. \cite{draganov2024shape} investigates the shape of words and their embeddings in Indo-European languages and find similar conclusions to Port et al. \cite{port2018persistent,port2019topological}, which investigate the syntactic topological space of such languages. They find that TDA features represent historical facts, such that languages clustered closely together are similar or influenced by each other. 
These applications show how TDA can be used to reveal and confirm linguistic phenomena. 
However, we currently observe only 13 theoretical TDA works in NLP, compared to over 100 non-theoretical ones. 
Thus, we need more theoretical TDA approaches, as it is impractical
to achieve the depth and understanding of performance from a topological perspective without further investigations.
% As this is a relevant non-trivial technique, we believe the NLP and ML community will benefit greatly from the creation of more theoretical approaches for TDA.

\subsection{Methodological Challenges}

\noindent \textbf{Interpretability of TDA Results.}
Interpreting TDA features in NLP problems, given its non-intuitive nature is very challenging. This is evident in the fact that most TDA for explainability applications is mostly in Computer Vision \cite{saul2018machine}, where the structure is clearly apparent. Consequently, the interpretation of TDA features for text or speech data remains an open problem. There are currently two main tasks in this space - 
(1) explain model performance by interpreting TDA features extracted from the model; and
(2) explain model performance by using TDA to probe the prediction space 
or data. Either task requires a deeper understanding of TDA such that intuitive explanations can be used to tie topology to linguistic phenomena. 
Specifically, we need novel approaches that link TDA features to 
linguistic phenomena, for instance, disentangling $\beta_0$, and $\beta_1$'s representations to different properties of natural texts such as coherency, and writing style. This can be done either through visualizing the latent space of a model \cite{rathore2023topobert,yan2025explainablemapperchartingllm,samaga2026halluzig} or probing the latent space 
of models with TDA \cite{spannaus2024topological,xenopoulos2022gale,solunke2024mountaineer,yan2025explainablemapperchartingllm}. 
% we can link topological features to linguistic principles \cite{port2018persistent}.
%, interpreting TDA features can be achievable but still remains an open problem. 

\noindent \textbf{TDA for Understanding NLP Model Behavior.}
% Existing studies suggest that TDA is promising for interpreting behaviors of NLP models, however, it remains scattered across tasks, representations, and modeling settings. For instance, TDA has been applied to probe the behavior of language models and make them less opaque. Existing work spans representation geometry and alignment across modalities \cite{haimmeirombobrowski2022unsupervised,sun2023topological,yessenbayev2022comparison,yessenbayev2024use,fay2025holes,vu2025topological,youtopological}, probing hidden states and model weights \cite{chauhan2022bertops,kudriashov2024more,fitz2024hidden}, training and optimization dynamics \cite{rathore2023topobert,garcia2024relative,topologyloss2025,more2025optimizing,zhang2026learning}, model-level explanation and diagnostics \cite{spannaus2024topological,gourgouliasestimating,yan2025explainablemapperchartingllm}, model efficiency \cite{proskura2024beyond,gardinazzi2024persistent,balderas2025green}, and safety, reliability, and reasoning-related behavior \cite{zadrozny2021abstraction,fitz2023large,kostenok2023uncertainty,samaga2026halluzig,varadarajan2025augmentingbiasdetectionllms,feng2024geometry,rair2025annotators,tan2025shape,li2025understanding,more2025optimizing,zhang2026learning,ishimtsevhole}. These studies, most of which are very recent, point to a broader open problem: how can TDA move beyond case-specific analysis to become a principled framework for interpreting complex model behavior in NLP?
Existing studies suggest that TDA is promising for interpreting the behavior of NLP models; however, current efforts remain scattered across tasks, representations, and model settings. In particular, TDA has been used to probe neural representations and make language models less opaque by characterizing the geometry and topology of their learned hidden space. For example, existing work spans representation geometry and alignment across modalities, including text embeddings, sentence semantics, text--speech alignment, adversarial versus non-adversarial latent spaces, and image--text alignment \cite{haimmeirombobrowski2022unsupervised,sun2023topological,yessenbayev2022comparison,yessenbayev2024use,fay2025holes,vu2025topological,youtopological, tsai2026shape}. Other studies use TDA to probe hidden states and model weights, revealing structural properties of language model latent spaces and uncovering grammatical or representational patterns \cite{chauhan2022bertops,kudriashov2024more,fitz2024hidden}. TDA has also been applied to training and optimization dynamics, model-level explanation and diagnostics, model efficiency \& safety, reliability, and reasoning-related behavior \cite{rathore2023topobert,garcia2024relative,topologyloss2025,more2025optimizing,zhang2026learning,spannaus2024topological,gourgouliasestimating,yan2025explainablemapperchartingllm,proskura2024beyond,gardinazzi2024persistent,balderas2025green,zadrozny2021abstraction,fitz2023large,kostenok2023uncertainty,samaga2026halluzig,varadarajan2025augmentingbiasdetectionllms,feng2024geometry,rair2025annotators,tan2025shape,li2025understanding,ishimtsevhole}. Beyond these works, how can TDA move beyond case-specific analysis to become a principled and generalizable framework for interpreting complex model behavior in NLP is of great value for future work.

\noindent \textbf{Improved TDA Feature Extraction and Representation Selection.}
Unlike some other data modalities that possess an intrinsic 
geometric structure, texts acquire their ``shape'' only 
through their numerical representations. In other words, the topology we observe is not an inherent property of the text 
itself, but of the embedding method used to encode it. A corpus represented with TF–IDF will therefore exhibit a 
geometry characteristic of sparse lexical weighting, whereas 
Transformer-based pre-trained embeddings induce a different 
shape driven by contextual semantic structure.
These differences are partly explained by the distinct 
linguistic features each representation captures - lexical 
frequency in the case of TF–IDF, and richer semantic and 
syntactic information in contextual embeddings. 
However, the resulting geometric and topological 
variations are often unintuitive. 
As a result, it becomes difficult to determine 
which numerical representation is most appropriate for 
extracting meaningful TDA features for a given task.
To address this challenge, we must develop principled methods for selecting representations that align with the objectives of the analysis. This includes exploring new forms of numerical encoding, such as \textit{symbolic representations} that capture the diversity of textual phenomena across tasks. 
Equally important is the development of improved strategies for leveraging existing embeddings in ways that enhance the stability, interpretability, and task-relevance of the extracted topological features.

% % Need for novel numerical representation 
% % Need to find better ways to use the representations we have 
% Unlike some other data modalities that have a distinct shape, 
% texts take the shape of their numerical representation. 
% This means that the shape of texts represented using TF-IDF will 
% embody the TF-IDF representation, which tends to be different from other 
% embeddings, such as the Transformer Pre-trained embeddings. 
% While the difference can be supported by the different linguistic features they capture, such as semantic, syntactic, and lexical features, the differences in shape can be unintuitive, making it difficult to determine the best numerical representation to extract the most useful TDA features. 
% Therefore, we must find novel ways to ascertain the appropriate numerical representation (based on the task) to extract topological features from.  
% Additionally, we need more numerical representation to capture the diversity in texts and tasks such as, the use of \textit{symbolic representations}. 
% Lastly, we need better ways to use the numerical representations that exist in such a way that is advantageous for extracting the best TDA features. 

\subsection{Frontier Expansions}

\noindent \textbf{Adversarial Robustness of TDA Features.}
Robustness to noise, particularly adversarial perturbations, has been an important research topic in NLP. While such robustness of TDA features is promising, there have been only a few works in this direction~\cite{perez2022topological,chauhan2022bertops,fay2025holes,vu2025topological}. For instance, Perez et al.~\cite{perez2022topological} shows that their \\topologically-augmented BERT model is more robust than the vanilla BERT model when tested against perturbations generated by TextAttack~\cite{morris2020textattack}. 
Chauhan et al. \cite{chauhan2022bertops} also show that there are some weak correlations between persistent homology features of a trained BERT model and its adversarial robustness against several state-of-the-art attackers. 
% Nevertheless, some results confirm that TDA is more robust, while others were inconclusive. 
Recently, we have observed interesting applications that use topology to track the latent space of LLMs before and after adversarial perturbations are introduced \cite{fay2025holes,vu2025topological}.
These studies highlight the emerging application of topology in robustness evaluations, which will benefit the NLP, ML, and security communities.

\noindent \textbf{Novel Applications of TDA.}
When we have more theoretical approaches of TDA and 
issues barring the application of TDA on interpretable NLP tasks are mitigated, we can hope that TDA can be applied to even more novel, diverse, and important tasks. 
From Section \ref{sec:tasks}, we can see that TDA has been applied to seven non-theoretical NLP tasks. 
While many of the tasks are interesting, especially the speech processing and health applications, there are still nuanced niche fields that could benefit from TDA. One glaring application is on multi-lingual tasks \cite{haimmeirombobrowski2022unsupervised,garcia2024relative,youtopological}. Due to the benefits of TDA, which include performing robustly on heterogeneous, imbalanced, and noisy data, its application to multi-lingual tasks is necessary. 
Other applications include: 
\textit{Topology-aware neural networks,
Topological interpretability,
semantic and syntactic structural analysis, 
forensic authorship, 
multi-modal (e.g., VLMs) analysis, }etc.

\noindent \textbf{Topological Deep Learning for NLP. } 
Due to the benefits of TDA and deep learning, a new niche field is born - 
Topological Deep Learning (TDL) as ``the collection of ideas and methods related to the use of topological concepts in deep learning'' \cite{papamarkou2024position}. 
Initially, TDL is described as an ensemble of topological features extracted by TDA techniques such as persistent homology and deep learning features. 
In this setting, TDL is a traditional deep learning model with extra features (i.e., TDA-extracted features). 
However, as the field has advanced, a new definition for TDL has emerged - ``the collection of ideas and methods related to the use of topological concepts in deep learning'' \cite{papamarkou2024position}. 
TDL allows a deep learning model to be integrated more deeply with concepts of algebraic topology, such as the introduction of simplicial neural networks (NNs) \cite{ebli2020simplicial,paluzo2024trainable}, which are NNs with layers made up of simplicial complexes. This deeper integration of TDA into NNs makes TDL particularly useful for the explosion of high-dimensional data.
% we are currently experiencing, especially because these datasets are often high-dimensional. 
These high-dimensional data require better tools for processing as the current tools shrink the dimension, resulting in information loss. In NLP, one particular approach to integrate TDA with high-dimensional NLP embeddings has been the utilization of text in graphical forms, which have been shown to yield better results than directly using texts as a sequence of tokens~\cite{liu2023coco,phan2023fake,li2025understanding,kushnareva2021artificial}. 
Nevertheless, more research is still needed to validate such an approach.
% to a variety of NLP tasks and problems.
% This introduces new and interdisciplinary research problems and opportunities to advance NLP further.
% Thus, as the niche field TDL grows, we must find a way to apply this novel concept on text or speech data to gain robust performance on classification tasks. 

\section{Conclusion}
Our world is currently experiencing an explosion of data and 
an explosion of computational techniques to process such data. 
Machine Learning (ML) is the most popular of these computational methods.
However, while its benefits are numerous, it has a few limitations. 
The biggest of the limitations of ML is its inability to sufficiently process 
data that is high-dimensional, imbalanced, noisy, and scarce. 
Therefore, a small community of NLP researchers emerged to tackle this limitation by proposing using TDA to tackle difficult NLP tasks. 
These researchers employ two TDA techniques - Persistent Homology
and Mapper to solve NLP tasks using theoretical and non-theoretical approaches. This yielded 137 papers, which we comprehensively surveyed in this paper. 
Finally, we conclude that while the applications of TDA in NLP have improved greatly since 2012, there is still room for improvement, specifically in reducing the barrier to entry for non-TDA experts to apply it to their NLP tasks.

\section{Ethical Statement}
This survey highlights emerging applications of Topological Data Analysis (TDA) in Natural Language Processing. While our primary goal is to synthesize existing work, we recognize that several use cases carry important ethical considerations and dual-use risks. 
Topological methods can inadvertently expose latent sensitive attributes (e.g., dialect, health cues, authorship), enabling re-identification or profiling even when data is anonymized. 
% Techniques discussed, such as watermarking, could also be repurposed to circumvent provenance systems. 
Applications in speech, emotion, and health-related domains further raise fairness, consent, and equity concerns, particularly for minority groups and low-resource languages.
Therefore, we emphasize the need for bias and robustness audits, careful data governance and licensing, and privacy-preserving mechanisms when sharing derived features. Responsible release practices, such as restricting code that enables circumvention, conducting red-team evaluations, and requiring IRB or ethics review for clinical or surveillance-adjacent uses, are essential. Finally, given the computational demands of TDA pipelines, their environmental impact should be considered as well.

\section{Acknowledgments}
The authors thank Dr. Charlie Dagli for his encouragement, reading the paper drafts, and providing invaluable recommendations.

\bibliographystyle{abbrv}
\bibliography{sigproc}  % sigproc.bib is the name of the Bibliography in this case

\appendix

\begin{table*}%[!htb]
    \centering
    \tiny
        \caption{Non-theoretical applications of TDA in NLP.
    For the TDA techniques - \textcolor{cyan}{PH}: Persistent Homology and \textcolor{orange}{M}: Mapper.
    Task categories - 
    \textcolor{green}{cl}: classification, 
    \textcolor{blue}{C \& TM}: clustering \& topic modeling, 
    \textcolor{brown}{S \& SA}: sentiment \& semantic analysis, 
    \textcolor{magenta}{S \& V}: structure \& visualization, 
    \textcolor{purple}{H,S,\& SA}: health, social, \& scholarly analysis, 
    \textcolor{teal}{S}: speech processing, 
    \textcolor{pink}{MI \& A}: model interpretation \& analysis (\textbf{PART I})}.
    % \footnotesize
    % \resizebox{9cm}{!}{
    \begin{tabular}{p{0.12\linewidth} p{0.07\linewidth}
    p{0.21\linewidth} p{0.15\linewidth} p{0.03\linewidth} p{0.02\linewidth} p{0.03\linewidth}}
    % \begin{tabular}{|c|c|c|c|c|c|}
    \toprule
    \multicolumn{1}{c}{\textbf{Name}}
    & \multicolumn{1}{c}{\textbf{Task}}
    & \multicolumn{1}{c}{\textbf{Problem}}
    & \multicolumn{1}{c}{\textbf{Numerical  Representation}}  % numerical representation
    & \multicolumn{1}{c}{\textbf{Learning Type}} % Supervised or unsupervised
    & \multicolumn{1}{c}{\textbf{Modality}}
    & \multicolumn{1}{c}{\textbf{TDA}}
    \\
    \midrule

    SIFT \cite{zhu2013persistent} & \textcolor{green}{cl} &
    child vs. adolescent writing detection & 
    TF-IDF & Supervised & Text & \textcolor{cyan}{PH}   \\
    \hdashline
    
    \cite{lovlie2023text}  &  \textcolor{green}{cl} &
    deepfake text detection & TF-IDF, GloVe & Supervised & Text & 
    \textcolor{cyan}{PH}     \\
    \hdashline

    \cite{huang2022tda} & \textcolor{green}{cl} &
    election speech feature extraction & 
    TF-IDF & Supervised & Text & 
    \textcolor{cyan}{PH}     \\
    \hdashline

    \cite{doshi2018movie,shin2019genre} & \textcolor{green}{cl} & movie genre & TF-IDF & Supervised & Text & 
    \textcolor{cyan}{PH}   \\
    \hdashline

    % \cite{shin2019genre} & movie genre \textcolor{green}{cl} & TF-IDF & Supervised &
    % Persistent Homology \\
    % \hline

    \cite{sovdat2016text} & \textcolor{green}{cl} & distinguishing between languages & 
    TF-IDF & Supervised & Text & \textcolor{cyan}{PH}   \\
    \hdashline

    \cite{savle2019topological} &\textcolor{brown}{S \& SA} & legal document entailment & TF-IDF & Supervised & Text & 
    \textcolor{cyan}{PH}   \\
    \hdashline

    \cite{michel2017does} &\textcolor{brown}{S \& SA} & clustering and sentiment analysis & TF-IDF, GloVe & Supervised & Text & 
    \textcolor{cyan}{PH}   \\
    \hdashline

    DoCollapse \cite{guan2016topological} & \textcolor{blue}{C \& TM} & keyphrase extraction & TF-IDF & Unsupervised & Text & 
    \textcolor{cyan}{PH}   \\
    \hdashline

    TOPOL \cite{torres2015topic} & \textcolor{blue}{C \& TM} & Twitter topic detection & 
    TF-IDF  & Supervised & Text & \textcolor{cyan}{PH}   \\
    \hdashline

    \cite{kumar2022extractive} & \textcolor{blue}{C \& TM} & 
    text summarization & TF-IDF & 
    Unsupervised & Text & 
    \textcolor{cyan}{PH}  \\
    \hdashline

    \cite{van2020novel} & \textcolor{green}{cl} & Propaganda tweet & TF-IDF & 
    Unsupervised & Text & \textcolor{orange}{M} \\
    \hdashline

    \cite{elyasi2019introduction} & \textcolor{green}{cl} & classify Persian poems & TF-IDF & Supervised & Text & 
    \textcolor{cyan}{PH}  \& \textcolor{orange}{M}  \\
    \hdashline

    \cite{effahtopological} & \textcolor{green}{cl} & age group categorization of lonely people & TF-IDF & Supervised & Text & 
    \textcolor{cyan}{PH}  \& \textcolor{orange}{M} \\
    \hdashline

    \cite{maadarani2020shape} & \textcolor{green}{cl} & nursery rhyme classification from different
    continents - Australia, Asia, Africa, Europe, and North America & TF-IDF & 
    Supervised & Text & 
    \textcolor{cyan}{PH}  \\
    \hline

    \cite{haghighatkhah2022story} & \textcolor{magenta}{S \& V} & building document structure &
    Pre-trained (Word2Vec) & Unsupervised & Text & 
    \textcolor{cyan}{PH}  \\
    \hdashline

    BERT+TDA \cite{wu2022topological} &\textcolor{brown}{S \& SA} & contradiction detection & 
    Pre-trained (Word2Vec) & Supervised & Text & 
    \textcolor{cyan}{PH}  \\
    \hdashline

    \cite{yessenbayev2022comparison} & \textcolor{pink}{MI \& A} &
    speaker recognition \& text processing & 
     Pre-trained (Word2Vec) & Unsupervised & Speech & \textcolor{cyan}{PH}  \\
    \hdashline

    \cite{yessenbayev2024use} & \textcolor{pink}{MI \& A} & speaker recognition \& text processing & 
     Pre-trained (Word2Vec) & Unsupervised & Speech & \textcolor{cyan}{PH}  \\
    \hdashline

    \cite{Filco306} &\textcolor{brown}{S \& SA} & building a topological search engine & 
    Pre-trained (Word2Vec) & Unsupervised & Text & 
    \textcolor{orange}{M} \\
    \hdashline

    \cite{holmes2020topological} & \textcolor{blue}{C \& TM} & document clustering and topic modeling tasks & 
    Pre-trained (Word2Vec) & Supervised & Text & 
    \textcolor{orange}{M} \\
    \hdashline

    \cite{rawson2022topological} &\textcolor{brown}{S \& SA} & word sense induction and disambiguation & 
    Pre-trained (Word2Vec) & Unsupervised & Text & 
    \textcolor{cyan}{PH}  \\
    \hdashline

    \cite{temvcinas2018local} &\textcolor{brown}{S \& SA} & word sense induction and disambiguation & 
    Pre-trained (Word2Vec, GloVe) & Unsupervised & Text & 
    \textcolor{cyan}{PH}  \\
    \hdashline

    \cite{feng2024geometry} & \textcolor{pink}{MI \& A} & Geometry of textual data augmentation &
    Pre-trained (Word2Vec) & Supervised & Text & \textcolor{cyan}{PH}  \\

    \hdashline

    \cite{tymochko2021connections} & \textcolor{green}{cl} & fraudulent paper detection & 
    Pre-trained (Word2Vec, GloVe, ElMo)  
    & Supervised & Text & 
    \textcolor{cyan}{PH}  \\
    \hdashline

    \cite{lordgrilo} & \textcolor{purple}{H,S,\& SA} & disease epidemic prediction & 
    Pre-trained (Word2Vec) & Supervised & Text & 
    \textcolor{cyan}{PH}  \\
    \hdashline

    \cite{yadav2025missing} & \textcolor{purple}{H,S,\& SA} & publication analysis & 
    Pre-trained (Word2Vec) & Unsupervised & Text & 
    \textcolor{cyan}{PH} \\

    \hdashline

    \cite{tymochko2020argumentative} & \textcolor{green}{cl} & finding topological loops in logical statements &
    Pre-trained (Word2Vec, GloVe) & Unsupervised & Text & 
    \textcolor{cyan}{PH}  \\
    \hdashline

    \cite{wright2020topological} & \textcolor{blue}{C \& TM} & distinguish subsets in data & 
    Pre-trained (Word2Vec) & Unsupervised & Text & 
    \textcolor{cyan}{PH}  \\
    \hdashline    

    \cite{paluzo2019towards} &\textcolor{brown}{S \& SA} & measuring the distance between the literary style of Spanish poets  &  Pre-trained (Word2Vec) & Supervised & Text & 
    \textcolor{cyan}{PH}  \\
    \hdashline

    \cite{bailey2025detecting} &\textcolor{brown}{S \& SA} & detecting narrative shifts & 
     Pre-trained (Word2Vec) & Unsupervised & Text & 
    \textcolor{cyan}{PH}  \\
    \hdashline

    \cite{alimpiev2020plot}  & \textcolor{brown}{S \& SA} & distinguishing news articles \& poems & Pre-trained (Word2Vec, GloVe) & Unsupervised & Text & 
    \textcolor{cyan}{PH}  \\
    \hline

    \cite{gholizadeh2018topological} & \textcolor{green}{cl} & extract the topological signatures of novelists & Pre-trained (GloVe) & Supervised & Text & 
    \textcolor{cyan}{PH}  \\
    \hdashline

    TIES \cite{gholizadeh2020novel} &\textcolor{brown}{S \& SA} & document categorization \& sentiment analysis & 
    Pre-trained (GloVe) & Unsupervised & Text & 
    \textcolor{cyan}{PH}  \\
    \hdashline

    \cite{spannaus2024topological} & \textcolor{pink}{MI \& A} & phenotype prediction and news group categorization & Pre-trained (GloVe) & Unsupervised & Text & 
    \textcolor{orange}{M} \\
    \hdashline

    \cite{zadrozny2021note} &\textcolor{brown}{S \& SA} & finding topological loops in logical statements &
    Pre-trained (GloVe) & Unsupervised & Text & 
    \textcolor{cyan}{PH}  \\
    \hdashline    

    \cite{byers2021hidden} & \textcolor{purple}{H,S,\& SA} & social anxiety detection & Pre-trained (GloVe) & Supervised & Text & 
    \textcolor{cyan}{PH} \\
    \hdashline 

    \cite{haimmeirombobrowski2022unsupervised} & \textcolor{pink}{MI \& A} & compare cross-lingual sentence representations & 
    Pre-trained (GloVe) & Unsupervised & Text & 
    \textcolor{cyan}{PH} \\
    \hdashline

    \cite{alanis2025detection} & \textcolor{green}{cl} & deepfake text detection & 
    Pre-trained (GloVe) & Supervised & Text & 
    \textcolor{cyan}{PH} \\
    \hdashline

    \cite{zadrozny2021abstraction} & \textcolor{pink}{MI \& A} & investigates the manifestations of intelligence and understanding in neural networks &  Pre-trained (GloVe) & Supervised & Text & \textcolor{cyan}{PH} \\
    \hdashline

    \cite{deng2022topological} & \textcolor{green}{cl} & fake news detection & Pre-trained (GloVe, BERT) & Supervised & Text & 
    \textcolor{cyan}{PH} \\
    \hdashline

    \cite{novak2019network} & \textcolor{purple}{H,S,\& SA} & analyzing scholarly network & Pre-trained (GloVe, BERT) & Unsupervised & Text & 
    \textcolor{cyan}{PH} \\
    \hline

    \cite{jakubowskietal2020topology} &\textcolor{brown}{S \& SA} & (1) polysemy word classification, and
    (2) word sense induction \& disambiguation & 
    Pre-trained (FastText) & Unsupervised & Text & 
    \textcolor{cyan}{PH} \\
    \hdashline

    \cite{triki2021analysis,shehu2024analysis} & \textcolor{brown}{S \& SA} &  polysemy word &
    Pre-trained (FastText) & Supervised & Text & 
    \textcolor{cyan}{PH} \\
    \hline

     PHD \cite{tulchinskii2024intrinsic} & \textcolor{green}{cl} & deepfake text detection & Pre-trained (Transformers - CLS)  
     & Supervised & Text & 
     \textcolor{cyan}{PH} \\
    \hdashline

    Short-PHD \cite{wei2025short} & \textcolor{green}{cl} & deepfake text detection (for short-text) & Pre-trained (Transformers - CLS)  
     & Supervised & Text & 
     \textcolor{cyan}{PH} \\
    \hdashline

    \cite{guilinger2025topological} & \textcolor{green}{cl} & deepfake text detection (for academic abstracts) & Pre-trained (Transformers - CLS)  
     & Supervised & Text & 
     \textcolor{cyan}{PH} \\
    \hdashline

    \cite{kushnareva2023artificial} & \textcolor{green}{cl} & deepfake text detection & Pre-trained (Transformers - CLS) & Supervised & Text & 
    \textcolor{cyan}{PH} \\
    \hdashline

    \cite{gourgouliasestimating} & \textcolor{pink}{MI \& A} & class separability estimation & 
    Pre-trained (Transformers - CLS) & Unsupervised & Text & 
    \textcolor{cyan}{PH} \\
    \hdashline

    TopoBERT \cite{rathore2023topobert} & 
     \textcolor{magenta}{S \& V} and \textcolor{pink}{MI \& A} &    
    visually analyzing the fine-tuning process of a Transformer-based model & Pre-trained (Transformers - CLS) &  Unsupervised & Text & \textcolor{orange}{M} \\
    \hdashline

    \cite{lavery2024combining} & \textcolor{green}{cl} & fake news detection & 
    Pre-trained (Transformers - CLS) &  Supervised & Text & \textcolor{cyan}{PH} \\
    \hdashline

    \cite{das2021persistence} & \textcolor{green}{cl} & classification of public speaking ratings from TED talks & Pre-trained (Transformers - CLS) &  Supervised & Text & \textcolor{cyan}{PH} \\
    \hdashline

     \cite{chandra2026topology} & 
    \textcolor{purple}{H,S,\& SA} and \textcolor{magenta}{S \& V} &
    tracking individual mental health journeys & Pre-trained (Transformers - cls) & Supervised & Text & \textcolor{orange}{M} \\
    \hdashline

    \cite{byrne2022topic,hopp2025persistent} & \textcolor{blue}{C \& TM} & topic modeling & Pre-trained (Transformers - CLS) &  Unsupervised & Text & \textcolor{orange}{M} \\
    \hdashline

    TOPFORMER \cite{uchendu2023topformer} & \textcolor{green}{cl} & deepfake text detection 
    & Pre-trained (Transformers - Hidden) & Supervised & Text & \textcolor{cyan}{PH} \\
    \hdashline

    TDA-BERTa \cite{rejimoan2025detection}  & \textcolor{green}{cl} & deepfake text detection 
    & Pre-trained (Transformers - Hidden) & Supervised & Text & \textcolor{cyan}{PH} \\
    \hdashline

    \cite{asriani2025nmt} & \textcolor{green}{cl} & Portuguese-English language translation 
    & Pre-trained (Transformers - Hidden) & Supervised & Text & \textcolor{cyan}{PH} \\

    % \cite{lavery2024combining} & \textcolor{green}{cl} & fake news detection & 
    % Pre-trained (Transformers - CLS) &  Supervised & Text & \textcolor{cyan}{PH} \\
    % \hdashline

    % \cite{bergomi2015dynamical} & 
    % \textcolor{teal}{S} & 
    % music & 
    % Multi-Modal & Supervised & Speech & \textcolor{cyan}{PH} \\
    % % \hline
    
    \bottomrule    
    \end{tabular}
    % }
    % }
    \label{tab:nontheory_label}
\end{table*}

\begin{table*}%[!htb]
    \centering
    \tiny
     \caption{Non-theoretical applications of TDA in NLP.
    For the TDA techniques - \textcolor{cyan}{PH}: Persistent Homology and \textcolor{orange}{M}: Mapper.
    Task categories - 
    \textcolor{green}{cl}: classification, 
    \textcolor{blue}{C \& TM}: clustering \& topic modeling, 
    \textcolor{brown}{S \& SA}: sentiment \& semantic analysis, 
    \textcolor{magenta}{S \& V}: structure \& visualization, 
    \textcolor{purple}{H,S,\& SA}: health, social, \& scholarly analysis, 
    \textcolor{teal}{S}: speech processing, 
    \textcolor{pink}{MI \& A}: model interpretation \& analysis  (\textbf{PART II})
    }
    % \resizebox{9cm}{!}{
    \begin{tabular}{p{0.12\linewidth} p{0.07\linewidth}
    p{0.21\linewidth} p{0.15\linewidth} p{0.03\linewidth} p{0.02\linewidth} p{0.03\linewidth}}
    % \begin{tabular}{|c|c|c|c|c|c|}
    \toprule
    \multicolumn{1}{c}{\textbf{Name}}
    & \multicolumn{1}{c}{\textbf{Task}}
    & \multicolumn{1}{c}{\textbf{Problem}}
    & \multicolumn{1}{c}{\textbf{Numerical  Representation}}  % numerical representation
    & \multicolumn{1}{c}{\textbf{Learning Type}} % Supervised or unsupervised
    & \multicolumn{1}{c}{\textbf{Modality}}
    & \multicolumn{1}{c}{\textbf{TDA}}
    \\
    \midrule
    % PHD \cite{tulchinskii2024intrinsic} & \textcolor{green}{cl} & deepfake text detection & Pre-trained (Transformers - CLS)  
    %  & Supervised & Text & 
    %  \textcolor{cyan}{PH} \\
    % \hdashline

    % Short-PHD \cite{wei2025short} & \textcolor{green}{cl} & deepfake text detection (for short-text) & Pre-trained (Transformers - CLS)  
    %  & Supervised & Text & 
    %  \textcolor{cyan}{PH} \\
    % \hdashline

    % \cite{guilinger2025topological} & \textcolor{green}{cl} & deepfake text detection (for academic abstracts) & Pre-trained (Transformers - CLS)  
    %  & Supervised & Text & 
    %  \textcolor{cyan}{PH} \\
    % \hdashline

    % \cite{kushnareva2023artificial} & \textcolor{green}{cl} & deepfake text detection & Pre-trained (Transformers - CLS) & Supervised & Text & 
    % \textcolor{cyan}{PH} \\
    % \hdashline

    % \cite{gourgouliasestimating} & \textcolor{pink}{MI \& A} & class separability estimation & 
    % Pre-trained (Transformers - CLS) & Unsupervised & Text & 
    % \textcolor{cyan}{PH} \\
    % \hdashline

    % TopoBERT \cite{rathore2023topobert} & 
    %  \textcolor{magenta}{S \& V} and \textcolor{pink}{MI \& A} &    
    % visually analyzing the fine-tuning process of a Transformer-based model & Pre-trained (Transformers - CLS) &  Unsupervised & Text & \textcolor{orange}{M} \\
    % \hdashline

    \cite{garcia2022applications} &\textcolor{brown}{S \& SA} & polysemy word & Pre-trained (Transformers - Hidden) & Supervised & Text & \textcolor{orange}{M} \\
    \hdashline

    \cite{bensalem2025detecting} & \textcolor{brown}{S \& SA} & polysemy word & Pre-trained (Transformers - Hidden) & Unsupervised & Text & \textcolor{cyan}{PH} \\
    \hdashline

    \cite{goshevtopology} & \textcolor{brown}{S \& SA} & semantic analysis of embeddings & Pre-trained (Transformers - Hidden) & Unsupervised & Text & \textcolor{cyan}{PH} \\
    \hdashline

    \cite{zhang2026text} & \textcolor{brown}{S \& SA} & text summarization & Pre-trained (Transformers - Hidden) & Unsupervised & Text & \textcolor{cyan}{PH} \\
    \hdashline
    
    \cite{alexander2023topological} & 
    \textcolor{purple}{H,S,\& SA} and \textcolor{magenta}{S \& V} &
    Hate speech, Misinformation \& Psychiatric disorder  & Pre-trained (Transformers - Hidden) & Supervised & Text & \textcolor{orange}{M} \\
    \hdashline

    PBCE \cite{balderas2025green} & \textcolor{pink}{MI \& A} &  Model compression & 
    Pre-trained (Transformers - Hidden) & Unsupervised & Text & 
    \textcolor{cyan}{PH} \\
    \hdashline
    
    Persistent Similarity \cite{gardinazzi2024persistent} & \textcolor{pink}{MI \& A} & 
    Probing layers in LLMs & 
    Pre-trained (Transformers - Hidden) & Unsupervised & Text & 
    \textcolor{cyan}{PH} \\
    \hdashline

    \cite{sun2023topological} & \textcolor{pink}{MI \& A} & 
    Correlation between sentence vectors & 
    Pre-trained (Word2Vec, Transformers - Hidden) & Unsupervised & Text & 
    \textcolor{cyan}{PH} \\
    \hdashline

    Persistence Scoring Function \cite{chauhan2022bertops} & 
    \textcolor{pink}{MI \& A} & 
     captures the homology of the high-dimensional hidden representations
     & Pre-trained (Transformers - Hidden) & Unsupervised & Text & 
    \textcolor{cyan}{PH} \\
    \hdashline

    Topological Densification \cite{garcia2024relative} & 
    \textcolor{pink}{MI \& A} & 
    zero-shot model stitching & Pre-trained (Transformers - Hidden) & Unsupervised & Text & 
    \textcolor{cyan}{PH} \\
    \hdashline

    TSLoss \cite{topologyloss2025} & 
     \textcolor{pink}{MI \& A} & 
    topology loss function for prompt tuning & Pre-trained (Transformers - Hidden) & Unsupervised & Text & 
    \textcolor{cyan}{PH} \\
    \hdashline

    HOLE \cite{athreya2025hole} & 
     \textcolor{pink}{MI \& A} & 
    Homological Observation of Latent Embeddings for Neural Network Interpretability
 & Pre-trained (Transformers - Hidden) & Supervised & Text & 
    \textcolor{cyan}{PH} \\
    \hdashline

    \cite{fitz2023large} & 
    \textcolor{pink}{MI \& A} & 
    topology of fairness & Pre-trained (Transformers - Hidden) & Unsupervised & Text & 
    \textcolor{orange}{M} \\
    \hdashline

    \cite{fay2025holes} & \textcolor{pink}{MI \& A} & adversarial vs. non-adversarial text representation in LLMs & Pre-trained (Transformers - Hidden) & Unsupervised & Text & 
    \textcolor{cyan}{PH} \\
    \hdashline

    \cite{tan2025shape} & \textcolor{pink}{MI \& A} & shape of reasoning process of LLMs & Pre-trained (Transformers - Hidden) & Unsupervised & Text & 
    \textcolor{cyan}{PH} \\
    \hdashline

    \cite{li2025understanding,zhang2026tda} & \textcolor{pink}{MI \& A} & reasoning process of LLMs using chain-of-thought prompts & Pre-trained (Transformers - Hidden) & Unsupervised & Text & 
    \textcolor{cyan}{PH} \\
    \hdashline

    EDTR \cite{more2025optimizing} & \textcolor{pink}{MI \& A} & measures LLM confidence & Pre-trained (Transformers - Hidden) & Unsupervised & Text & 
    \textcolor{cyan}{PH} \\
    \hdashline

    GHS-TDA \cite{zhang2026learning} & \textcolor{pink}{MI \& A} & improving reasoning process of chain-of-thought & Pre-trained (Transformers - Hidden) & Unsupervised & Text & 
    \textcolor{cyan}{PH} \\
    \hdashline

    \cite{ishimtsevhole} & \textcolor{pink}{MI \& A} & tracing reasoning process of chain-of-thought & Pre-trained (Transformers - Hidden) & Unsupervised & Text & 
    \textcolor{cyan}{PH} \\
    \hdashline
    
    \cite{ruppik2024local} & \textcolor{blue}{C \& TM} & dialogue term extraction & Pre-trained (Transformers - Hidden) & Supervised & Text & \textcolor{cyan}{PH} \\
    \hdashline

    \cite{kudriashov2024more} & \textcolor{pink}{MI \& A} & polypersonality & Pre-trained (Transformers - Hidden) & Unsupervised & Text & \textcolor{cyan}{PH} \\
    \hdashline

    \cite{yan2025explainablemapperchartingllm} & \textcolor{pink}{MI \& A} & explainability of latent space & Pre-trained (Transformers - Hidden) & Unsupervised & Text & \textcolor{cyan}{PH} \\
    \hdashline

    \cite{fitz2024hidden} & \textcolor{pink}{MI \& A} & topological complexity of LLM hidden space & Pre-trained (Transformers - Hidden) & Unsupervised & Text & \textcolor{cyan}{PH} \\
    \hdashline

    \cite{bazarova2025hallucination,voronkova2026topology} & \textcolor{green}{cl} & LLM Hallucination detection &
    Pre-trained (Transformers - Attention) & Supervised & Text & \textcolor{cyan}{PH} \\
    \hdashline

    \cite{macdonald2024binary} & \textcolor{green}{cl} & Code attribution &
    Pre-trained (Transformers - Attention) & Supervised & Text & \textcolor{cyan}{PH} \\
    \hdashline

    \cite{kushnareva2021artificial} & \textcolor{green}{cl} & deepfake text detection & 
    Pre-trained (Transformers - Attention) & Supervised & Text & \textcolor{cyan}{PH} \\
    \hdashline

    \cite{cherniavskii2022acceptability,proskurina2023can,jain2024beyond} & 
   \textcolor{brown}{S \& SA} &
    grammatical acceptability judgment & 
    Pre-trained (Transformers - Attention) & Supervised & Text & \textcolor{cyan}{PH} \\
    \hdashline

    \cite{kostenok2023uncertainty} & 
    \textcolor{pink}{MI \& A} & 
    Uncertainty estimation of model predictions &
    Pre-trained (Transformers - Attention) & Supervised & Text & \textcolor{cyan}{PH} \\
    \hdashline

    \cite{samaga2026halluzig} & 
    \textcolor{pink}{MI \& A} & 
    hallucination detection in LLMs &
    Pre-trained (Transformers - Attention) & Supervised & Text & \textcolor{cyan}{PH} \\
    \hdashline

    \cite{tsai2026shape} & 
    \textcolor{pink}{MI \& A} & 
    investigation of adversarial influence in the latent space &
    Pre-trained (Transformers - Attention) & Supervised & Text & \textcolor{cyan}{PH} \\
    \hdashline

    \cite{varadarajan2025augmentingbiasdetectionllms} & 
    \textcolor{pink}{MI \& A} & 
    bias detection &
    Pre-trained (Transformers - Attention) & Supervised & Text & \textcolor{cyan}{PH} \\
    \hdashline

    \cite{perez2022topological} & \textcolor{green}{cl} & 
    spam detection, grammatical acceptability judgment, and movie sentiment analysis & 
    Pre-trained (Transformers - Attention) & Supervised & Text & \textcolor{cyan}{PH} \\
    \hdashline

    \cite{sakurai2025authorship} & \textcolor{green}{cl} & Authorship attribution of Japanese texts &
     Pre-trained (Transformers - Attention) & Supervised & Text & \textcolor{cyan}{PH} \\
    \hdashline

    \cite{pollano2023detecting} & \textcolor{green}{cl} & out-of-distribution detection & Pre-trained (Transformers - Attention, CLS) & Supervised & Text & \textcolor{cyan}{PH} \\
    \hdashline

    \cite{proskura2024beyond} & \textcolor{pink}{MI \& A} & estimation of weights for ensembles of classification models & Pre-trained (Transformers - Attention, CLS) & Supervised & Text & \textcolor{cyan}{PH} \\
    \hdashline

    \cite{rair2025annotators} &  \textcolor{pink}{MI \& A} & visualizes fine-tuning process to see when annotators disagree & Pre-trained (Transformers - Attention, CLS) & Supervised & Text & \textcolor{cyan}{PH} \\
    \hdashline

    \cite{arun2025topo} & \textcolor{magenta}{S \& V} and \textcolor{brown}{S \& SA} & 
    controversial vs. non-controversial political discourse detection & Pre-trained (Transformers - CLS) & Supervised & Text & \textcolor{cyan}{PH} \\
    \hdashline

    PHPS \cite{meng2024encoding} & \textcolor{magenta}{S \& V} and \textcolor{brown}{S \& SA} & 
    personalized web search & Pre-trained (Transformers - CLS) & Supervised & Text & \textcolor{cyan}{PH} \\
    \hdashline

    \cite{snopov2024vulnerability} & \textcolor{green}{cl} & vulnerability detection in code 
    & Pre-trained (Transformers - Attention) & Supervised & Text & \textcolor{cyan}{PH} \\
    \hdashline

    TopoHuBERT \cite{tulchinskii2023topological} & \textcolor{teal}{S} & 
    speaker recognition & Pre-trained (Transformers - Attention) & Supervised & Speech & \textcolor{cyan}{PH} \\
    \hdashline

    \cite{vukovic2022dialogue} &\textcolor{brown}{S \& SA} &
    dialogue term extraction & Pre-trained (Transformers - Attention) & Supervised & Text & \textcolor{cyan}{PH} \\
    \hline

    \cite{yurchuk2023tongue} & 
   \textcolor{brown}{S \& SA} & 
    Ukrainian tongue twisters & 
    Symbolic Representations &  Supervised & Text & \textcolor{cyan}{PH} \\
    \hdashline

    \cite{kovaliuk2024topological} & 
    \textcolor{teal}{S} & 
    Ukrainian tongue twisters & 
    Symbolic Representations &  Supervised & Speech & \textcolor{cyan}{PH} \\
    \hdashline

    \cite{escolar2025topological} & 
    \textcolor{purple}{H,S,\& SA} & 
    recipe discovery & 
    Symbolic Representations &  Unsupervised & text & 
    \textcolor{cyan}{PH} \\
    
    \hline
    \cite{bonafos2023topological} & 
    \textcolor{teal}{S} &
    human vowel & 
    Multi-Modal & Supervised & Speech & \textcolor{cyan}{PH} \\
    \hdashline
    
    \cite{bonafos2024dirichlet} & 
    \textcolor{teal}{S} & 
    infant vocalization  & 
    Multi-Modal & Unsupervised & Speech & \textcolor{cyan}{PH} \\
    \hdashline

    \cite{yu2025topologicaldeeplearningspeech,lameris2024topological} & 
    \textcolor{teal}{S} & 
    speech recognition  & 
    Multi-Modal & Unsupervised & Speech & \textcolor{cyan}{PH} \\
    \hdashline

    \cite{gonzalez2019towards,paluzo2022emotion,raskin2025shape} & 
    \textcolor{teal}{S} & 
    emotion recognition & 
    Multi-Modal & Supervised & Speech & \textcolor{cyan}{PH} \\
    \hdashline

    \cite{tlachac2020topological} & 
    \textcolor{teal}{S} & 
    depression detection & 
    Multi-Modal & Supervised & Speech & \textcolor{cyan}{PH} \\
    \hdashline

    \cite{zhu2024topology} & 
    \textcolor{teal}{S} & 
    consonants recognition & 
    Multi-Modal & Supervised & Speech & \textcolor{cyan}{PH} \\
    \hdashline

    % \cite{reise2024topological} & song recognition & 
    % Time Series & Supervised & Audio & Persistent Homology \\
    % \hline

    %  TopFusion \cite{myers2023topfusion} & image \& audio \textcolor{green}{cl} & 
    % Time Series & Supervised & Audio & Persistent Homology \\
    % \hline

    %  \cite{sassone2022bridging} & 
    %  \textcolor{teal}{S} & 
    %  music genre & 
    % Multi-Modal & Supervised & Speech & \textcolor{cyan}{PH} \\
    % \hdashline

    \cite{vu2025topological} & 
    \textcolor{pink}{MI \& A} & 
    multi-modal adversarial robustness assessment  & 
    Multi-Modal & Supervised & Text & \textcolor{cyan}{PH} \\
    \hdashline

    \cite{huang2025topology,zhang2024homology,rahim2024topological,xu2026topo} & 
    \textcolor{pink}{MI \& A} & 
     aligning the topological structures of image and text representations during tuning  & 
    Multi-Modal & Supervised & Text & \textcolor{cyan}{PH} \\
    \hdashline

     \cite{bachiri2025topological} & 
    \textcolor{pink}{MI \& A} & 
     aligning the topological structures of image and text representations during tuning  & 
    Multi-Modal & Supervised & Text & \textcolor{cyan}{PH} \\
    \hdashline

    \cite{youtopological}  & 
    \textcolor{pink}{MI \& A} & 
    multi-modal multi-lingual topological alignment  & 
    Multi-Modal & Supervised & Text & \textcolor{cyan}{PH} \\
    % \hdashline

    % \cite{brown2009nonlinear} & vowels, nasals, and noise \textcolor{green}{cl} & 
    % Time Series & Supervised & Audio & Persistent Homology \\
    % \hline

    %  \cite{liu2016applying} & music auto-tagging & 
    % Time Series & Supervised & Audio & Persistent Homology \\
    % \hline

    % \cite{emrani2014real} & wheezing detection & 
    % Time Series & Supervised & Audio & Persistent Homology \\
    % \hline

    % \cite{bergomi2015dynamical} & 
    % \textcolor{teal}{S} & 
    % music \textcolor{green}{cl} & 
    % Multi-Modal & Supervised & Speech & \textcolor{cyan}{PH} \\
    % % \hline
    
    \bottomrule    
    \end{tabular}
    % }
   
    \label{tab:nontheory_labelb}
\end{table*}

\end{document}